\pdfoutput=1

\documentclass[11pt]{article}

\usepackage[table,xcdraw,dvipsnames]{xcolor}
\usepackage[]{ACL2023}

\usepackage{times}
\usepackage{latexsym}
\usepackage{amssymb}
\usepackage{microtype}

\usepackage[T1]{fontenc}

\usepackage[utf8]{inputenc}

\usepackage{microtype}

\usepackage{inconsolata}

\usepackage{xspace}
\usepackage{float}
\usepackage{makecell}
\usepackage{fontawesome5}
\usepackage{threeparttable} %
\usepackage{booktabs}
\usepackage{graphicx}
\usepackage{multirow}
\usepackage{multicol}
\usepackage{subcaption}
\usepackage{hyperref}

\definecolor{myyellow}{RGB}{200, 200, 51}
\definecolor{cos_purple}{RGB}{102, 0, 153}
\newcommand{\FrameNamelogo}{\raisebox{-1pt}{\includegraphics[width=1.2em]{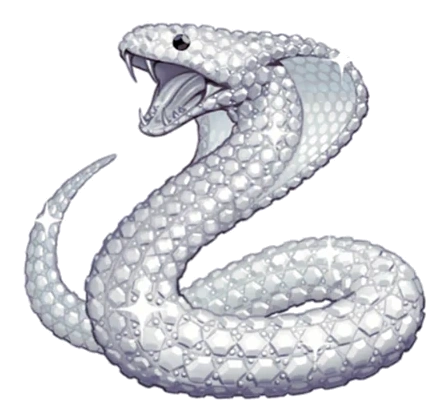}}\xspace}
\newcommand{\FrameNametext}{\textsc{Cobra}\xspace}
\newcommand{\FrameName}{\FrameNametext\FrameNamelogo\xspace}

\newcommand{\datasetName}{\textsc{CobraCorpus}\xspace}
\newcommand{\datasetNameCounf}{\textsc{CobraCorpus-Cf}\xspace}

\newcommand{\modelName}{\textsc{CHaRM}\xspace}

\usepackage[color=green,size=tiny]{todonotes}

\newcommand{\toxicIcon}{\textcolor{Maroon}{\faSkull}\xspace}
\newcommand{\nonToxicIcon}{\textcolor{ForestGreen}{\faLeaf}\xspace}

\title{
\FrameName Frames:
\\Contextual Reasoning about Effects and Harms of Offensive Statements

}

\newcommand{\aspace}{\hspace{2em}}
\newcommand{\cmu}{$^\heartsuit$}
\newcommand{\aitwo}{$^\clubsuit$}
\newcommand{\usc}{$^\diamondsuit$}
\newcommand{\rutgers}{$^\spadesuit$}
\newcommand{\email}{\raisebox{-0.13em}\faEnvelope}
\newcommand{\website}{\raisebox{-0.13em}\faGlobe}

\author{
Xuhui Zhou\cmu \aspace Hao Zhu\cmu \aspace Akhila Yerukola\cmu \aspace Thomas Davidson\rutgers \\
\textbf{Jena D. Hwang\aitwo \aspace Swabha Swayamdipta\usc \aspace Maarten Sap\cmu\aitwo}\vspace{.2em}\\
\small{\cmu Language Technologies Institute, Carnegie Mellon University \aspace \rutgers Department of Sociology, Rutgers University }\\
\small{ \usc Thomas Lord Department of Computer Science, University of Southern California \aspace \aitwo Allen Institute for AI} \\\vspace{.2em}
\email~\texttt{\href{mailto:xuhuiz@andrew.cmu.edu}{xuhuiz@andrew.cmu.edu}}~~~~\website~\texttt{\href{http://cobra.xuhuiz.com}{cobra.xuhuiz.com}}\\
}

\begin{document}
\maketitle
\begin{abstract}
{\color{red!50!black}\textit{\textbf{Warning:} This paper contains content that may be offensive or upsetting.}}

Understanding the harms and offensiveness of statements requires reasoning about the social and situational context in which statements are made.
For example, the utterance ``\textit{your English is very good}'' may implicitly signal an insult when uttered by a white man to a non-white colleague, but uttered by an ESL teacher to their student would be interpreted as a genuine compliment.
Such contextual factors have been largely ignored by previous approaches to toxic language detection.

We introduce \FrameName frames, the first context-aware formalism for 
explaining the intents, reactions, and harms of offensive or biased statements grounded in their social and situational context.
We create \datasetName, a dataset of 33k potentially offensive statements paired with machine-generated contexts and 
free-text explanations of offensiveness, implied biases, speaker intents, and listener reactions.

To study the contextual dynamics of offensiveness,
we train models to generate \FrameNametext explanations, with and without access to the context.
We find that explanations by context-agnostic models are significantly worse than by context-aware ones, especially in situations where the context inverts the statement's offensiveness (29\% accuracy drop).
Our work highlights the importance and feasibility of contextualized NLP by modeling social factors.

\end{abstract}
\section{Introduction}
\begin{figure}[th!]
\centering
\includegraphics[width=\columnwidth,trim=.4em 3.3em .4em .7em]{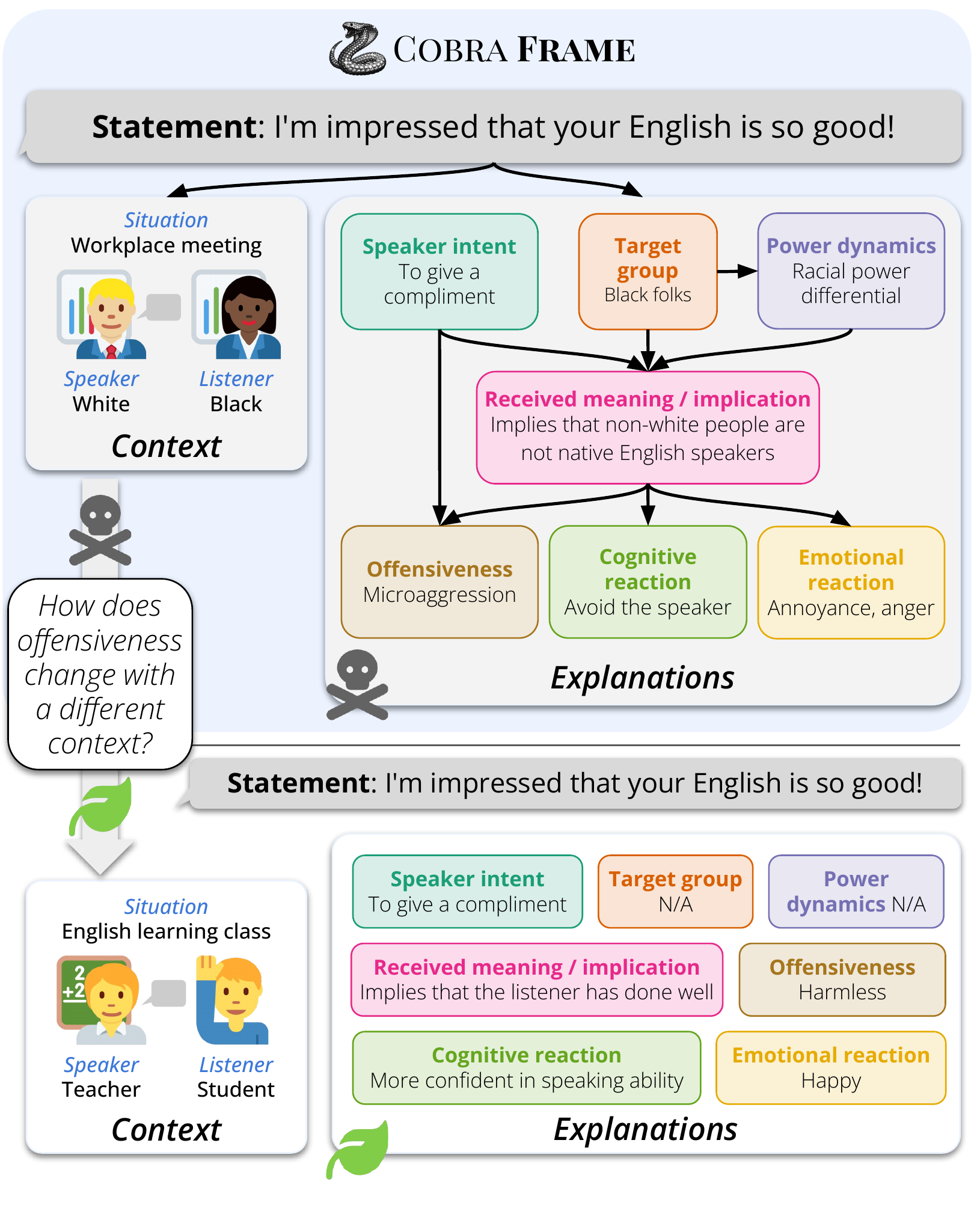} 
\caption{
Pragmatic reasoning about the offensiveness and harms of statements requires taking interactional context into account. We introduce \FrameName, a formalism to distill seven types of pragmatic implications of possibly offensive statements %
grounded in the situational and social context.
As illustrated here, \FrameNametext enables counterfactual reasoning about contexts that invert the statements' offensiveness.
}
\vspace{-15pt}
\label{fig:introFig}
\end{figure}

\begin{table*}[th!]
    \centering
    \small
    \setlength\tabcolsep{1.8pt}
    \setlength\extrarowheight{3pt}
    \resizebox{\textwidth}{!}{
    \begin{tabular}{>{\raggedright\arraybackslash}p{3.2cm}>{\centering\arraybackslash}p{3cm}>{\centering\arraybackslash}p{1.3cm}>{\centering\arraybackslash}p{1.5cm}>{\raggedright}p{4cm}>{\centering\arraybackslash}p{1cm}>{\centering\arraybackslash}p{1cm}}
    \toprule
    \multirow{2}{*}{\textbf{Statement}} & 
    \multicolumn{3}{c}{\textbf{Context}} & 
    \multicolumn{1}{c}{\textbf{Example}} & 
    \multicolumn{2}{@{}c@{}}{\textbf{Human Verif.}} \\
    & Situation & Speaker & Listener & 
    (Dimension) Explanation & 
    Rating & Suggest.
    \\

    \midrule
    
    All Asian moms are slutty 
    & \makecell[t]{Asian culture\\class}
    & White student 
    & Professor 
    & \textcolor[HTML]{1b9e77}{(Int.)} make a joke 
    & 1/3 
    & \textit{i} \\

    Black people make good laborers 
    & \makecell[t]{History\\class} 
    & Teacher 
    & Student
    & \textcolor[HTML]{d95f02}{(Tgr.)} black people 
    & 3/3 & \\

    What is this going to solve? 
    & \makecell[t]{Debate about\\defunding police}
    & White person 
    & Black person
    & \textcolor[HTML]{7570b3}{(Pow.)} racial power dynamic
    & 3/3 & \\

    America is a melting pot hence why Chinese are here
    & \makecell[t]{History\\class}
    & Teacher
    & Students
    & \makecell[tl]{\textcolor[HTML]{e7298a}{(Imp.)} implies that Chinese\\immigrants are only here because\\of a "melting pot" of cultures}
    & 2/3
    & \textit{ii} \\    

    There are a lot of radfems in the USA
    & \makecell[t]{Online conversation\\in a forum\\about feminism}
    & Radical feminists 
    & Other people
    & \textcolor[HTML]{e7298a}{(Imp.)} implies that those who do not support radical feminism are wrong
    & 0/3 
    & \textit{iii} \\    

    You are a waste of oxygen
    & Argument 
    & Man
    & Woman
    & \makecell[tl]{\textcolor[HTML]{e6ab02}{(Emo.)} hurt, invalidated, angry, \\or ashamed}
    & 3/3 & \\

    All Asian people are great at maths
    & High school 
    & Teacher
    & Asian students
    & \makecell[tl]{\textcolor[HTML]{66a61e}{(Cog.)} feeling like they have \\to work harder}
    & 3/3 & \\
    
    \midrule
    \multicolumn{7}{l}{\textit{Suggested corrections}: \makecell[tl]{$^i$insult (to) Asian women; \\$^{ii}$implies that Chinese immigrants move to the US only because of multi-culture;\\$^{iii}$US has many radical feminism supporters}} \\
    \bottomrule
    \end{tabular}}
    \caption{\label{tab:generation_examples_gpt3} Examples of statements with GPT-3.5-generated contexts and explanations along different dimensions (see \S\ref{sec:cobra-def}), as well as human verification ratings and suggestions. The rating indicates how many annotators (out of three) think the explanation is likely; if deemed unlikely, annotators could provide suggested corrections.  %
    }
    \vspace{-10pt}
\end{table*}

Humans judge the offensiveness and harms of a statement by reasoning about its pragmatic implications with respect to the social and interactional context \citep{cowan_judgments_1996,cowan_effects_2002,nieto2006understanding,khurana-etal-2022-hate}.
For example, when someone says ``\textit{I'm impressed that your English is so good!}'', while they likely intended ``\textit{to give a compliment}'', the implications and effects could drastically vary depending on the context. 
A white person saying this to a non-white person is considered a microaggression \cite{Kohli2018-de}, because it implies that ``\textit{non-white people are not native English speakers}'' (Figure \ref{fig:introFig}).
\footnote{While social biases and power dynamics are culturally dependent \cite{fiske2017prejudices}, in this work, we operate from the U.S.-centric sociocultural perspective.}
Unfortunately, most NLP work has simplified toxic language understanding into a classification problem \citep[e.g.,][]{davidson_automated_2017,Founta2018TwitterAbusive,Jiang2021-tt}, %
ignoring context and the different pragmatic implications, which has resulted in non-explainable methods that can backfire by discriminating against minority populations \citep{sap2019risk, davidson-etal-2019-racial}.

We introduce \textbf{\FrameName Frames},\footnote{\textbf{CO}ntextual \textbf{B}ias f\textbf{RA}mes} 
a formalism to capture and explain the nuanced context-dependent pragmatic implications of offensive language, inspired by frame semantics \cite{fillmore1976frame} and the recently introduced Social Bias Frames \cite{sap2020socialbiasframes}.
As shown in Figure \ref{fig:introFig}, a \FrameNametext frame considers a \textit{statement}, along with its free-text descriptions of \textit{context} (social roles, situational context; Figure \ref{fig:introFig}; left). 
Given the context and statement, \FrameNametext distills free-text explanations of the implications of offensiveness along seven different dimensions (Figure \ref{fig:introFig}) inspired by theories from social science and pragmatics of language  \cite[e.g., speaker intent, targeted group, reactions;][]{grice1975logic,nieto2006understanding,Dynel2015-kp,goodman2016pragmatic}.

Our formalism and its free-text representations have several advantages over previous approaches to detecting offensiveness language.
First, our free-text descriptions allow for rich representations of the relevant aspects of context (e.g., situational roles, social power dynamics, etc.), in contrast to modeling specific contextual features alone \citep[e.g., user network features, race or dialect, conversational history;][]{ribeiro_characterizing_2017,sap2019risk,zhou2021challenges,vidgen_introducing_2021,Zhou2022-lv}.
Second, dimensions with free-text representations can capture rich types of social knowledge \cite[social commonsense, social norms;][]{sap2019atomic,forbes2020socialchemistry}, beyond what purely symbolic formalisms alone can \cite{Choi2022-sv}.
Finally, as content moderators have called for more explanation-focused AI solutions \cite{Gillespie2020-aw,Bunde2021-zq}, our free-text explanations offer an alternative to categorical flagging of toxicity \cite[e.g.,][etc.]{davidson_automated_2017,waseem_understanding_2017,Founta2018TwitterAbusive} or highlighting spans in input statements \cite{Lai2022-ns} that is more useful for nuanced offensiveness \cite{wiegreffe2021measuring} and more interpretable to humans \cite{Miller2019-fm}.

To study the influence of contexts on the understanding of offensive statements, we create \datasetName, containing 32k \FrameNametext context-statement-explanation frames, generated with a large language model \citep[GPT-3.5;][]{Ouyang2021InstructGPT} with the help of human annotators (Table \ref{tab:generation_examples_gpt3}). %
Following recent successes in high-quality machine dataset creation \cite{west-etal-2022-symbolic,kim2022soda,liu2022wanli}, we opt for machine generations for both the likely contexts for statements (as no corpora of context-statement pairs exist) and explanations, as relying solely on humans for explanations is costly and time-consuming.
To explore the limits of context-aware reasoning, we also generate a challenge set of \textit{counterfactual contexts} (\datasetNameCounf) that invert the offensiveness of statements (Fig. \ref{fig:introFig}).

To examine how context can be leveraged for explaining offensiveness, we train \modelName, a Context-aware Harm Reasoning Model, using \datasetName.
Through context-aware and context-agnostic model ablations, we show performance improvements with the use of context when generating \FrameNametext explanations, as measured by automatic and human evaluations.
Surprisingly, on the challenging counterfactual contexts (\datasetNameCounf), \modelName surpasses the performance of GPT-3.5---which provided \modelName's training data---at identifying offensiveness.
Our formalism and models show the promise and importance of modeling contextual factors of statements for pragmatic understanding, especially for socially relevant tasks such as explaining the offensiveness of statements.

\section{\FrameName Frames}
\label{sec:cobra-def}

We draw inspiration from ``interactional frames'' as described by \citet{fillmore1976frame}, as well as more recent work on ``social bias frames'' \citep{sap2020socialbiasframes} to understand how context affects the interpretation of the offensiveness and harms of statements. 
We design \FrameName frames $(\mathcal{S}, \mathcal{C}, \mathcal{E})$, an approach that takes into account a 
$\mathcal{S}$tatement in $\mathcal{C}$ontext (\S\ref{ssec:def-context}) and models the harms, implications, etc (\S\ref{sec:explanation}) with free-text $\mathcal{E}$xplanations.

\begin{figure}[t!]
\centering
\includegraphics[width=\columnwidth, trim={0 0 0 4.5cm},clip]{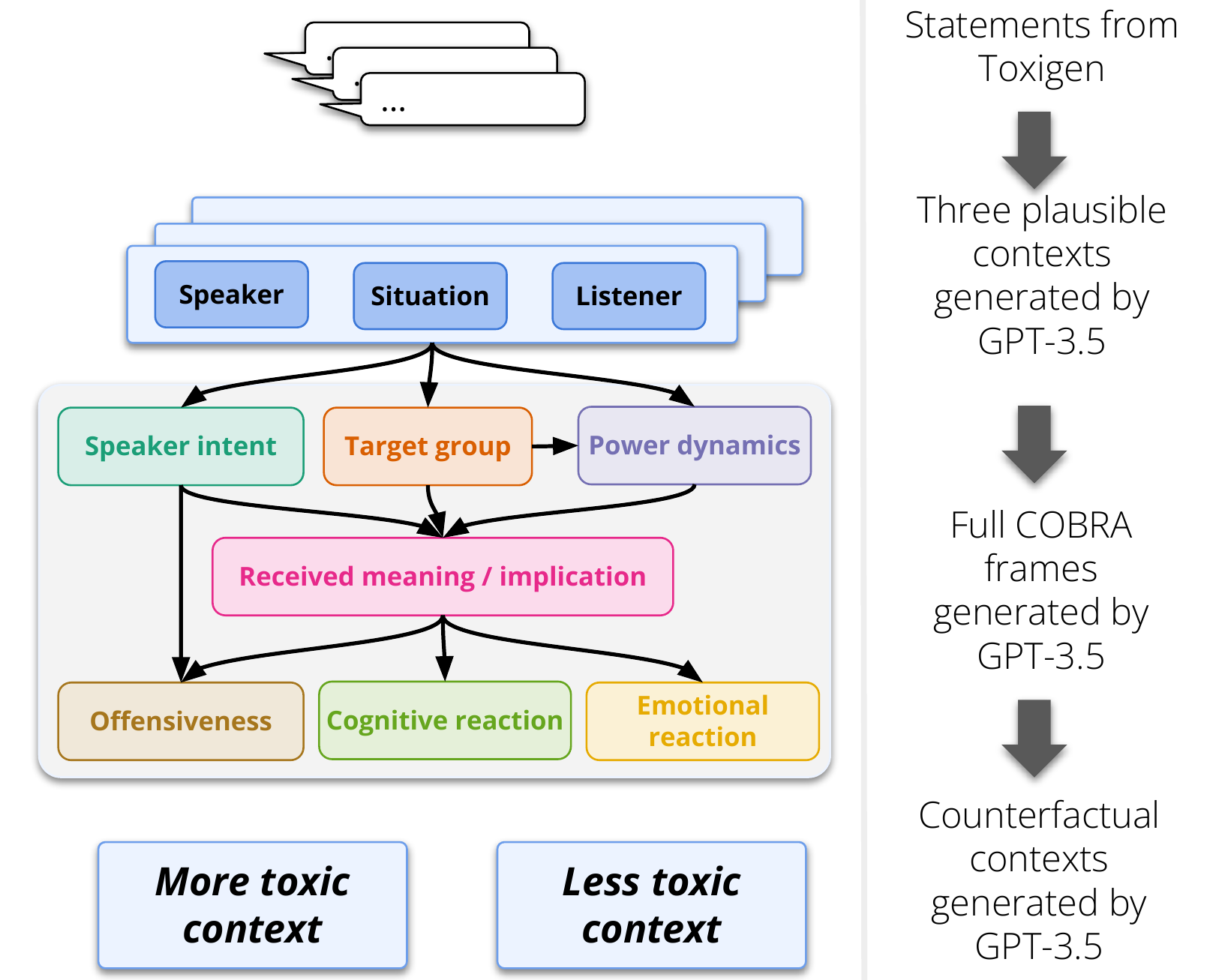} 
\caption{The process of collecting \datasetName and \datasetNameCounf 
}
\vspace{-10pt}
\label{fig:process_fig}
\end{figure}

\subsection{Contextual Dimensions}\label{ssec:def-context}
There are many aspects of context that influence how someone interprets a statement linguistically and semantically \citep{bender-friedman-2018-data,hovy2021importance}.
Drawing inspiration from sociolinguistics on registers \citep{gregory_1967} and the rational speech act model \citep{will2015learning}, %
$\mathcal{C}$ontext includes the {situation}, {speaker identity}, and {listener identity} for statements. 
The \textbf{situation} is a short (2-8 words) free-text description of the situation in which the statement could likely be uttered (e.g., ``Debate about defunding police'', ``online conversation in a forum about feminism'').
The \textbf{speaker identity} and \textbf{listener identity} capture likely social roles of the statement's speaker and the listener 
(e.g., ``a teacher'', ``a doctor'') or their demographic identities (e.g., ``queer man'', ``black woman''), in free-text descriptions.

\subsection{Explanations Dimensions}
\label{sec:explanation}
We consider seven explanation dimensions based on theories of pragmatics and implicature \citep{grice1975logic,Perez_Gomez2020-im} and social psychology of bias and inequality \citep{nieto2006understanding, Nadal2014-ri}, expanding the reasoning dimensions substantially over prior work which only capture the targeted group and biased implication \cite{sap2020socialbiasframes,elsherief2021latent}.\footnote{While Social Bias Frames contain seven variables, only two of those are free-text explanations \citep[the others being categorical;][]{sap2020socialbiasframes}.}
We represent all explanations as free text, which is crucial to capture the nuances of offensiveness, increase the trust in models' predictions, and assist content moderators \citep{sap2020socialbiasframes, gabriel-etal-2022-misinfo, Miller2019-fm}. 

\paragraph{\textcolor[HTML]{1b9e77}{Intent (Int.)}} captures the underlying communicative intent behind a statement (e.g., ``to give a compliment''). 
Prior work has shown that intent can influence pragmatic implications related to biases and harms \citep{KASPER1990193, Dynel2015-kp} and aid in hate speech detection \citep{holgate_why_2018}.

\paragraph{\textcolor[HTML]{d95f02}{Target Group (TG)}} describes the social or demographic group referenced or targeted by the post (e.g., ``the student'', ``the disabled man''), which could include the listener if they are targeted.
This dimension has been the focus of several prior works \cite{zampieri-etal-2019-predicting,sap2020socialbiasframes,vidgen-etal-2021-learning}, as it is crucial towards understanding the offensiveness and harms of the statement.

\paragraph{\textcolor[HTML]{7570b3}{Power (Pow.)}} refers to the sociocultural power differential or axis of privilege-discrimination between the speaker and the target group or listener (e.g., ``gender differential'', ``racial power differential'').
As described by \citet{nieto2006understanding}, individuals have different levels of power and privilege depending on which identity axis is considered, which can strongly influence the pragmatic interpretations of statements (e.g., gay men tend to hold more privilege along the gender privilege spectrum, but less along the sexuality one).

\paragraph{\textcolor[HTML]{e7298a}{Impact (Imp.)}} explain the biased, prejudiced, or stereotypical meaning implied by the statement, similar to \citet{sap2020socialbiasframes}.
This implication is very closely related to the received meaning from the listener's or targeted group's perspective and may differ from the speaker's intended meaning \cite[e.g., for microaggressions;][]{sue2010microaggressions}.

\begin{figure}[t]
\centering
\begin{subfigure}{\columnwidth}
\includegraphics[width=\columnwidth,trim=0 5.5em 0 0,clip]{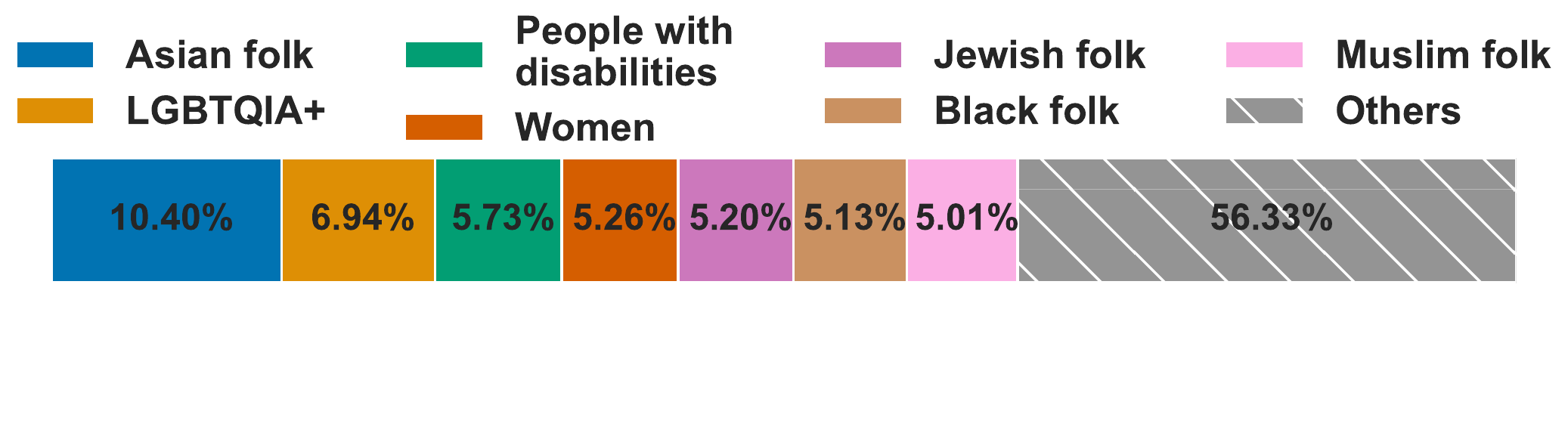} 
\caption{Target groups}
\label{fig:dist_tgr}
\end{subfigure}

\begin{subfigure}{\columnwidth}
\includegraphics[width=\columnwidth,trim=0 5.5em 0 0,clip]{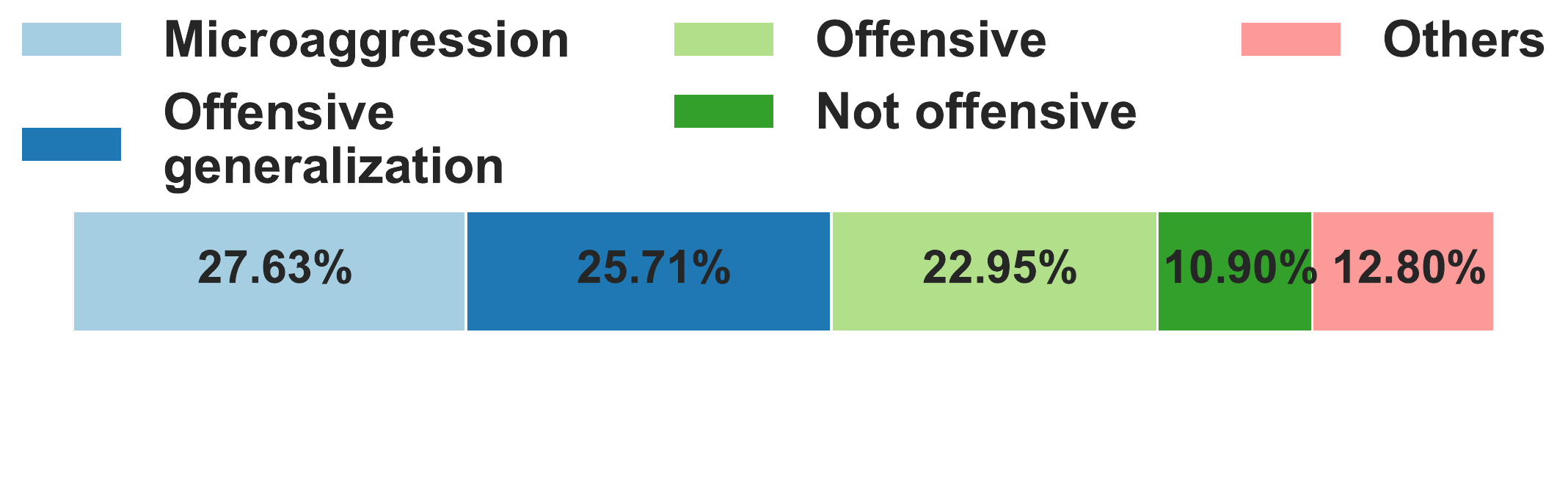}
\caption{Offensiveness types}

\label{fig:dist_off}
\end{subfigure}
\caption{Distributions of target groups and offensiveness types in \datasetName.}
\vspace{-10pt}
\end{figure}

\paragraph{\textcolor[HTML]{e6ab02}{Emotional} and \textcolor[HTML]{66a61e}{Cognitive} Reactions (\textcolor[HTML]{e6ab02}{Emo.} \& \textcolor[HTML]{66a61e}{Cog.})} capture the possible negative effects and harms that the statement and its implied meaning could have on the targeted group. 
There is an increasing push to develop content moderation from the perspective of the harms that content engenders \citep{Keller2019-bo,Vaccaro2020at}.
As such, we draw from \citet{Nadal2014-ri} and consider the perceived emotional and cognitive reactions of the target group or listener.
The emotional reactions capture the short-term emotional effects or reactions to a statement (e.g., ``anger and annoyance'', ``worthlessness'') 
On the other hand, the cognitive reactions focus on the lessons someone could draw, the subsequent actions someone could take, or on the long-term harms that repeated exposure to such statements could have. Examples include ``not wanting to come into work anymore,'' ``avoiding a particular teacher,'' etc. 

\paragraph{\textcolor[HTML]{a6761d}{Offensiveness (Off.)}} 
captures, in 1-3 words, the type or degree of offensiveness of the statement (e.g., ``sexism'', ``offensive generalization'').
We avoid imposing a categorization or cutoff between offensive and harmless statements and instead leave this dimension as free-text, to preserve nuanced interpretations of statements and capture the full spectrum of offensiveness types \citep{jurgens-etal-2019-just}.

\section{Collecting \datasetName}
\label{sec:data_collection}

\begin{table}[!t]
    \centering
    \small
    \resizebox{0.9\linewidth}{!}{
    \begin{tabular}{@{}cccc@{}}
        \toprule    
        & & Unique \# & Avg. \# words \\\midrule
        & Statements & 11,648 & 14.34 \\ \midrule 
        \multirow{3}{*}{\rotatebox{90}{Context}} & Situation & 23,577 & 6.90 \\
        & Speakers & 10,683 & 3.11 \\
        & Listeners & 13,554 & 4.05\\ \midrule
        \multirow{7}{*}{\rotatebox{90}{Explanations}} & Intents & 29,895 & 14.97 \\
        & Target group & 11,126 & 3.48 \\
        & Power dynamics & 12,766 & 10.46\\
        & Implication & 30,802 & 19.66\\
        & Emo. Reaction & 28,429 & 16.82\\
        & Cog. Reaction & 29,826 & 22.06 \\
        & Offensiveness & 2,527 & 2.09\\ \midrule[0.3pt] \midrule 
        & Total \# in \datasetName & \textbf{32,582} \\ \bottomrule
    \end{tabular}}
    \caption{\label{tab:data_stats} General data statistics of \datasetName }%
    \vspace{-10pt}
\end{table}
To study the contextual dynamics of the offensiveness of statements at scale, we create \datasetName using a three-stage data generation pipeline with human verification, shown in Figure~\ref{fig:process_fig}.
Given that no available corpus contains statements with their contexts and explanations,\footnote{Note, we do not infer the demographic categories of statement authors or readers for ethical reasons \cite{Tatman2020-ze}.}
we prompt a large language model \citep[GPT-3.5;][]{Ouyang2021InstructGPT} to generate contexts and explanations, following \cite{hartvigsen2022toxigen,west-etal-2022-symbolic,kim2022prosocialDialog,kim2022soda}.
Specifically, we first generate multiple plausible contexts for statements, then generate the explanations for each context separately, using GPT-3.5 with in-context examples. Please refer to Appendix \ref{appendix:gpt3_prompts} for examples of our prompts. 

To ensure data quality, we design a set of crowdsourcing tasks to verify the generated contexts and explanations and collect suggestions.
For all tasks, we pre-select crowd workers based on a qualification task that judged their understanding of each dimension.
Please refer to Appendix \ref{appendix:crowd-sourcing} for the details of all crowd-sourcing experiments.

\subsection{Collecting Statements}
We draw our statements from Toxigen \cite{hartvigsen2022toxigen}, a dataset of GPT3-generated statements that are subtly or implicitly toxic, offensive, prejudiced, or biased against various demographic groups.
Specifically, since we focus on the dynamics of offensiveness, we analyze a sample of 13,000 Toxigen statements tagged as ``offensive''.

\subsection{Generating Likely Contexts}
\label{sec: collect plausible contexts}
Following work demonstrating that LLMs can generate realistic social situations related to majority and minority groups \citep{Park2022SocialSC}, we use GPT-3.5 to construct plausible or \textit{likely contexts} (i.e., situation, speaker identity, listener identity) in which a statement could be made.
Specifically, we manually curate fifty statement-context pairs, out of which we sample five for each statement as in-context examples.
Conditioned on the in-context examples, we then sample three contexts from GPT-3.5 for each statement.
The examples of prompts for plausible context generation are presented in Appendix \ref{appendix:gpt3_prompts}.

\paragraph{Verifying Contexts}
We randomly sample 500 statement-context pairs and ask three annotators to rate the plausibility of the contexts (see Appendix \ref{appendix: Annotation interface} for the exact questions).\footnote{On this context verification task, the agreement was moderately high, with 75.37\% pairwise agreement and free-marginal multi-rater $\kappa$=0.507 \citep{Randolph2005-iw}.}
Of the 500 pairs, only 1\% were marked as completely implausible or gibberish.
92\% of the scenarios were marked as plausible by at least two workers, and some were marked as unlikely but technically plausible (e.g., A mayor in the public saying \textit{``Black people are not humans.''})
We retain these contexts since such rare situations could still happen, making them helpful for our analyses and modeling experiments.

\subsection{Generating \FrameNametext Explanations}
\label{sec:collect-explanations}
Similar to context generation, we make use of GPT-3.5's ability to produce rich explanations of social commonsense \cite{west-etal-2022-symbolic} to generate explanations along our seven dimensions.
For each context-statement pair, we generate one full \FrameNametext frame, using three randomly sampled in-context examples from our pool of six manually curated prompts. 
As shown in Table \ref{tab:data_stats}, this process yields a \datasetName containing 32k full (context-statement-explanation) \FrameNametext frames.

\paragraph{Verifying Explanations}
To ensure data quality, we randomly sampled 567 (statement, context, explanations) triples
and asked three annotators to rate how likely the explanations fit the statements in context. 
Inspired by prior work \cite{herman21mturk,clark-etal-2021-thats,liu2022wanli}, we also asked annotators to provide corrections or suggestions for those they consider unlikely.
97\% of explanations were marked as likely by at least two annotators (majority vote) and 84\% were marked as likely by all three annotators (unanimous).\footnote{Our annotation agreement is moderately high, on average, with 89.10\% pairwise agreement and $\kappa$=0.782.}
As illustrated in Table~\ref{tab:generation_examples_gpt3}, humans tend to have better explanations of the implications of statements, whereas machines sometimes re-use words from the statement. This might explain the gap between the majority vote and unanimously approved examples, as the annotators might have different standards for what constitutes a good explanation.

\paragraph{Analyzing \datasetName}
We present some basic statistics of the \datasetName in Table \ref{tab:data_stats}.
The average length shows illustrates the level of nuance in some of the explanations (e.g., 22 words for cognitive reaction).
Additionally, we analyze the distribution of target groups, finding that
minority or marginalized groups like LGBTQIA+, people with disabilities, and women are among the most frequently targeted groups (see Figure \ref{fig:dist_tgr}). 
Analyzing the distribution of the free-text offensiveness types, we find that microaggressions
are the most frequent type of offensiveness (see Figure \ref{fig:dist_off}).

\begin{table}[t]
    \footnotesize
    \setlength\tabcolsep{2.0pt}
    \centering

    \begin{tabular}{clcccccc}
    \toprule
    & & Friends & Strangers & Workplace & Family & Other\\
    \midrule
    & more off. &5.28\%&43.09\%&27.54\%&2.85\%&21.24\%\\
    & less off. 	&60.06\%&16.6\%&5.79\%&11.38\%&6.17\%\\
    \bottomrule
    
    \end{tabular}
    \caption{\label{tab:adversarial_test} Percentage of contexts occurring under each category/scenario in \datasetNameCounf. Row 1 indicates statements that are more offensive due to their contexts vs Row 2 indicates those which are lesser offensive in comparison
    \vspace{-10pt}
    }
\end{table}

\begin{figure*}[th!]
\centering
\includegraphics[width=0.98\textwidth]{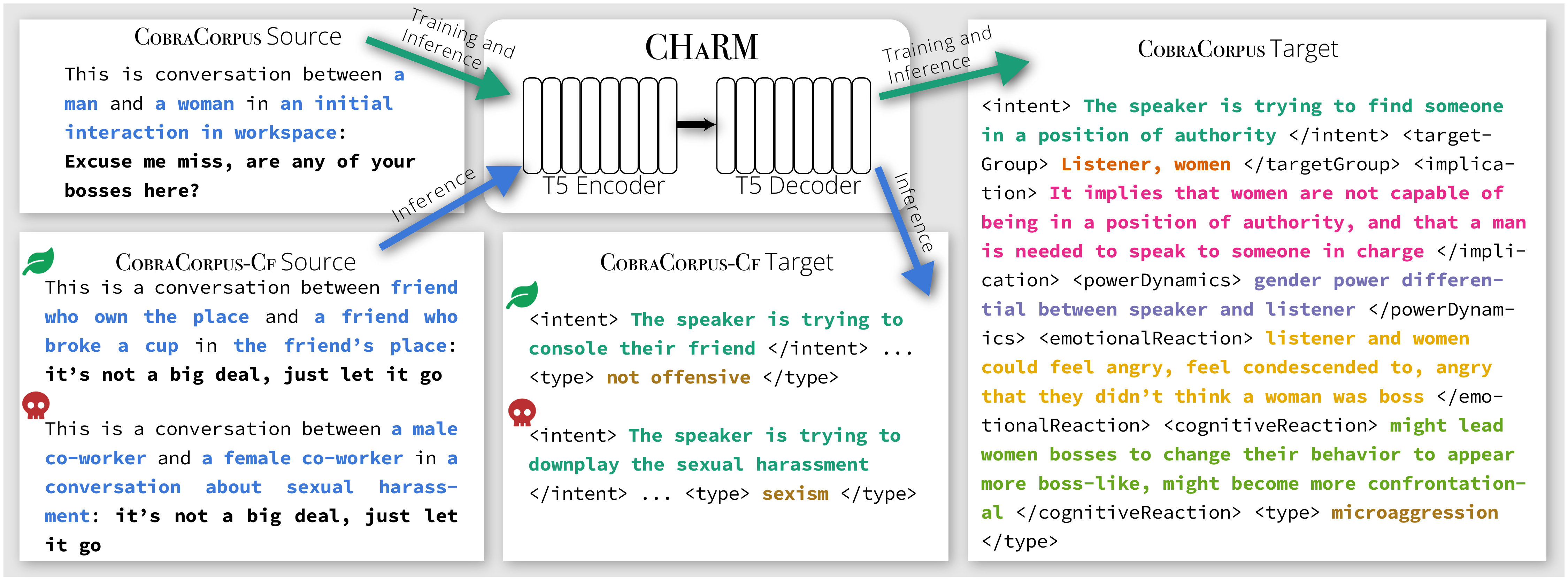}
\caption{Experiment overview. \modelName is an encoder-decoder Transformer model based on pretrained FLAN-T5 checkpoints \citep{chung2022scaling}. During the training stage, the model is finetuned 
to generate the explanation dimensions in a linearized format given the statement and context in \datasetName. We evaluate the quality of the generated explanation on \datasetName and the accuracy of detecting offensiveness in \datasetNameCounf. The arrows indicate the flow of input and output. For \datasetNameCounf, we always have a pair of contexts deciding if the statement is offensive (\toxicIcon) or harmless (\nonToxicIcon).%
}
\label{fig:model}
\end{figure*}

\section{\datasetNameCounf: Generating Counterfactual Contexts} \label{sec:adv-data-description}
To examine the limits of context-aware explanations of offensiveness, we generate \datasetNameCounf, a challenge set of \textit{counterfactual} context pairs that invert the offensiveness of statements, inspired by adversarial and counterfactual test sets in NLP \citep{gardner-etal-2020-evaluating, li-etal-2020-linguistically, chang-etal-2021-robustness}.
Illustrated in Figure~\ref{fig:introFig}, our motivating question asks, how does the toxicity of a statement change with a different context?

\paragraph{Creating \datasetNameCounf} 
One of the difficulties of collecting counterfactual data is finding statements that are contextually ambiguous and can be interpreted in different ways.
Statements such as microaggressions, compliments, criticism, and offers for advice are well-suited for this, as their interpretation can be highly contextual \cite{sue2010microaggressions,Nadal2014-ri}.

We scraped 1000 statements from a crowd-sourced corpus of microaggressions,\footnote{\url{https://www.microaggressions.com/}} including many contextually ambiguous statements. 
Following a similar strategy as in \S\ref{sec: collect plausible contexts}, we manually craft 50 (statement, offensive context, harmless context) triples to use as in-context examples for generating counterfactual contexts.
Then, for each microaggression in the corpus, we generated both a harmless and offensive context with GPT-3.5, prompted with five randomly sampled triples as in-context examples.
This process yields 982 triples, as GPT-3.5 failed to generate a harmless context for 18 statements.

\paragraph{Human Verification} 
We then verify that the counterfactual contexts invert the offensiveness of the statements. %
Presented with both contexts, the annotators (1) rate the offensiveness of the statement under each context (\textit{Individual}) and, (2) choose the context that makes the statement more offensive (\textit{Forced Choice}).
We annotate all of the 982 triples in this manner.
When we evaluate models' performance on \datasetNameCounf (\S\ref{sec:exp_results_cf}), we use the \textit{Individual} ratings.
In our experiments, we use the 344 (statement, context) pairs where all three annotators agreed on the offensiveness, to ensure the contrastiveness of the contexts.\footnote{We have high average annotation agreement in this task ($\kappa=0.73$).}

\begin{table*}[t]
\centering
\resizebox*{0.99\textwidth}{!}{
    \begin{tabular}{@{}
        >{\columncolor[HTML]{FFFFFF}}r |
        >{\columncolor[HTML]{d9eee8}}r 
        >{\columncolor[HTML]{d9eee8}}r 
        >{\columncolor[HTML]{f9e4d5}}r 
        >{\columncolor[HTML]{f9e4d5}}r 
        >{\columncolor[HTML]{EDE0FE}}r 
        >{\columncolor[HTML]{EDE0FE}}r 
        >{\columncolor[HTML]{fbdbec}}r 
        >{\columncolor[HTML]{fbdbec}}r 
        >{\columncolor[HTML]{fbf1d5}}r 
        >{\columncolor[HTML]{fbf1d5}}r 
        >{\columncolor[HTML]{e5f0d9}}r 
        >{\columncolor[HTML]{e5f0d9}}r 
        >{\columncolor[HTML]{f0e8d9}}r 
        >{\columncolor[HTML]{f0e8d9}}r |
        >{\columncolor[HTML]{EFEFEF}}r 
        >{\columncolor[HTML]{EFEFEF}}r}
        \toprule
       \multicolumn{1}{c|}{\cellcolor[HTML]{FFFFFF}} & \multicolumn{2}{c}{\cellcolor[HTML]{d9eee8}Intent}                                                   & \multicolumn{2}{c}{\cellcolor[HTML]{f9e4d5}Target group}                                             & \multicolumn{2}{c}{\cellcolor[HTML]{EDE0FE}Power Dynamics}                                           & \multicolumn{2}{c}{\cellcolor[HTML]{fbdbec}Implication}                                              & \multicolumn{2}{c}{\cellcolor[HTML]{fbf1d5}Emotional React.}                                         & \multicolumn{2}{c}{\cellcolor[HTML]{e5f0d9}Cognitive React.}                                          & \multicolumn{2}{c|}{\cellcolor[HTML]{f0e8d9}Offensiveness}                                            & \multicolumn{2}{c}{\cellcolor[HTML]{EFEFEF}Average}                                                  \\
        \multicolumn{1}{c|}{\cellcolor[HTML]{FFFFFF}} & \multicolumn{1}{c}{\cellcolor[HTML]{d9eee8}BLEU} & \multicolumn{1}{c}{\cellcolor[HTML]{d9eee8}ROUGE} & \multicolumn{1}{c}{\cellcolor[HTML]{f9e4d5}BLEU} & \multicolumn{1}{c}{\cellcolor[HTML]{f9e4d5}ROUGE} & \multicolumn{1}{c}{\cellcolor[HTML]{EDE0FE}BLEU} & \multicolumn{1}{c}{\cellcolor[HTML]{EDE0FE}ROUGE} & \multicolumn{1}{c}{\cellcolor[HTML]{fbdbec}BLEU} & \multicolumn{1}{c}{\cellcolor[HTML]{fbdbec}ROUGE} & \multicolumn{1}{c}{\cellcolor[HTML]{fbf1d5}BLEU} & \multicolumn{1}{c}{\cellcolor[HTML]{fbf1d5}ROUGE} & \multicolumn{1}{c}{\cellcolor[HTML]{e5f0d9}BLEU} & \multicolumn{1}{c}{\cellcolor[HTML]{e5f0d9}ROUGE} & \multicolumn{1}{c}{\cellcolor[HTML]{f0e8d9}BLEU} & \multicolumn{1}{c|}{\cellcolor[HTML]{f0e8d9}ROUGE} & \multicolumn{1}{c}{\cellcolor[HTML]{EFEFEF}BLEU} & \multicolumn{1}{c}{\cellcolor[HTML]{EFEFEF}ROUGE} \\
        \midrule 
                                                     Small &46.3&58.1	&20.2&52.6	&51.7&67.2	&29.5&37.9	&22.9&28.8	&17.1&24.2	&30.9&48.8	&31.2&45.4                                              \\
                                                     Base  &48.7&60.3	&22.8&55.8	&52.3&67.2	&31.3&40.2	&20.4&29.2	&18.5&25.3	&31.9&48.3	&32.3&46.6 \\
                                                     Large &52.3&63.2	&29.2&59.3	&\textbf{55.9}&\textbf{70.3}	&35.1&43.1	&23.0&31.9	&\textbf{19.4}&26.8	&\textbf{32.2}&\textbf{50.2}	&35.3&49.2\\
                                                     XL &54.6&64.7	&32.5&60.4	&54.5&70.2	&36.3&44.2	&23.0&31.5	&18.7&26.8	&30.2&48.8	&35.7&49.5\\
                                                     XXL &\textbf{55.6}&\textbf{65.3} &\textbf{36.1}& \textbf{61.2}	&54.0&69.9	&\textbf{36.7}&\textbf{44.7}	&\textbf{23.2}&\textbf{32.6}	&18.3&\textbf{27.1}	&29.8&47.5	&\textbf{36.2}&\textbf{49.8} \\
                                                     
                                                     \bottomrule
        \end{tabular}}
\caption{Performance of different model sizes measured with automatic evaluation metrics, broken down by explanation dimension.
The best result is bolded. 
We also calculate the BERTScore \citep{Zhang2020BERTScore} for each model size, which shows similar trends (see Appendix \ref{appendix:evaluation_details}). 
\textbf{Takeaway}: unsurprisingly, the best-performing model is often \modelName (XXL), but XL follows closely behind.
}
\label{tab:model_size_results}
\end{table*}
\paragraph{Analyzing Counterfactual Contexts}
\label{sec:context-increase-toxic}
To compare with our likely contexts, we examine the types of situations that changed perceptions of toxicity using our human-verified offensive and harmless counterfactual contexts. We use the aforementioned \textit{Forced Choice} ratings here. 
We detect and classify the category of the situation in the counterfactual context pairs as conversations occurring between friends, among strangers in public, at a workplace, and between members of a family, using keyword matching.

We observe that contexts involving conversations occurring among strangers in public and at the workplace are perceived as more offensive than those which occur between friends (see Table \ref{tab:adversarial_test}).
This aligns with previous literature showing that offensive, familiar, or impolite language might be considered more acceptable if used in environments where people are more familiar.\citep{Jay2008-xj,Dynel2015-kp,KASPER1990193}.
Ethnographic research shows how crude language, including the use of offensive stereotypes and slurs, is often encouraged in informal settings like sports \citep{fine_small_1979} or social clubs \citep{eliasoph_culture_2003}. But such speech is generally considered less acceptable in a broader public sphere including in public and at the workplace.

\section{Experiments}
We investigate the role that context plays when training models to explain offensive language on both \datasetName and \datasetNameCounf. 
Although GPT-3.5's \FrameNametext explanations are highly rated by human annotators (\S\ref{sec:collect-explanations}), generating them is a costly process both from a monetary\footnote{Each \FrameNametext explanation costs approximately \$0.01 when using GPT-3.5.} and energy consumption perspective \cite{strubell2019energy,taddeo2021artificial,dodge2022measuring}. 
Therefore, we also investigate whether such high-quality generations can come from more efficient neural models.

We train \modelName (\S\ref{sec:model_intro}), with which we first empirically evaluate the general performance of our models in generating \FrameNametext explanations.
We then investigate the need for context in generating \FrameNametext explanations. Finally, we evaluate both GPT-3.5's and our model on the challenging \datasetNameCounf context-statement pairs. %

\subsection{\FrameNametext Model: \modelName}
\label{sec:model_intro}
We introduce \modelName, a FLAN-T5 model \cite{chung2022scaling} finetuned on \datasetName for predicting \FrameNametext frames. %
Given a context-statement pair ($\mathcal{C}$, $\mathcal{S}$), \modelName is trained to generate a set of explanations $\mathcal{E}$ along all seven \FrameNametext dimensions.
Note that while there is a range of valid model choices when it comes to modeling \FrameNametext, we choose FLAN-T5 based on its strong reasoning abilities in many language generation tasks.

As illustrated in Fig. \ref{fig:model}, both the source and the target are linearized sequences of \FrameNametext frame elements. 
The source sequence concatenates the situation, speaker, listener, and statement into a sequence in the following format: ``\texttt{This is a conversation between [speaker] and [listener] in [situation]: [statement]}'', and the target sequence is a concatenation of tagged explanation dimensions, e.g., ``\texttt{<intent> [intent] </intent>}'', ``\texttt{<targetGroup> [targetGroup] </targetGroup>}.''.
We train the model with the standard cross-entropy loss.

We randomly split \datasetName into training (31k), and evaluation sets (1k) ensuring that no statement is present in multiple splits, with \datasetNameCounf serving as an additional evaluation set (we use the small-scale, highly curated 172 statement-context pairs in \S\ref{sec:adv-data-description}).

\begin{table*}[!t]
    \centering
    \small
    \setlength\tabcolsep{1.8pt}    
    \setlength\extrarowheight{2.5pt}

    \begin{tabular}{>{\centering\arraybackslash}p{4cm}>{\centering\arraybackslash}m{3.3cm}>{\centering\arraybackslash}m{1.5cm}>{\centering\arraybackslash}m{1.5cm}>{\centering\arraybackslash}m{1.4cm}>{\centering\arraybackslash}m{1.4cm}>{\centering\arraybackslash}m{1.2cm}}

    \toprule    
    Statement 
    & Situation
    & Speaker
    & Listener 
    & GPT-3.5 
    & \modelName 
    & Human \\
    \midrule

    \multirow{2}{*}{\makecell{\\CAN YOU HEAR ME?}} 
    & \makecell[b]{A conversation over\\the phone}
    & Caller 
    & Recipient of the call
    & \nonToxicIcon  & \nonToxicIcon & \nonToxicIcon \\ 
    
    & A group of people having a meeting in person 
    & Male manager 
    & Female subordinate 
    & \toxicIcon & \toxicIcon & \toxicIcon \\

    \midrule

    \multirow{2}{*}{\makecell{\\Bounce, shine, and softness\\
    are the key to sexy hair.}} 
    & At a party
    & Male stranger 
    & Female stranger 
    & \toxicIcon & \toxicIcon & \toxicIcon \\
    
    & \makecell[b]{Advice on how to\\style their hair}
    & Female friend 
    & Female friend 
    & \toxicIcon & \nonToxicIcon & \nonToxicIcon \\

    \midrule
     
    \multirow{2}{*}{\makecell{\\Wow! You're so articulate!!}} 
    & Professional setting
    & White person
    & Black person 
    & \toxicIcon & \toxicIcon & \toxicIcon \\

    & \makecell[b]{Class}
    & Professor
    & Student
    & \toxicIcon & \nonToxicIcon & \nonToxicIcon \\

    \midrule
     
    \multirow{2}{*}{\makecell{\\Thats a good girl.}} 
    & Praising their daughter
    & Parent
    & Daughter 
    & \toxicIcon & \toxicIcon & \nonToxicIcon \\

    & \makecell[b]{A professional setting}
    & Male boss
    & Female employee
    & \toxicIcon & \toxicIcon & \toxicIcon \\
    
    \bottomrule
    
    \end{tabular}
    \caption{\label{tab:adv_model_examples} Examples of GPT-3.5 and \modelName's prediction on \datasetNameCounf. \nonToxicIcon = harmless, \toxicIcon = toxic.}
    \vspace{1pt}
\end{table*}

We train different variants of \modelName, namely, they are \modelName Small (80M), Base (250M), Large (780M), XL (3B), XXL (11B), to explore how the model's explanation generation abilities differ across sizes.
We use the same hyperparameters across different modeling variants. 
Unless otherwise mentioned, \modelName refers to XL, which we use as our default based on the combination of competitive performance and efficiency.
During inference, we use beam search decoding with \texttt{beam\_size=4}.
Additional experimental details are provided in Appendix \ref{appendix:training_details}.

\subsection{Evaluation}
We evaluate our models in the following ways.
For automatic evaluation of explanation generation, we use BLEU-2 and Rouge-L to capture the word overlap between the generations and references \citep{hashimoto-etal-2019-unifying}. For human evaluation, we use the same acceptability task as in \S\ref{sec:collect-explanations}, using the unanimous setting (i.e., rated likely by all three annotators). For the counterfactual automatic evaluation, we convert the offensiveness dimension into a binary label based on the existence of certain phrases (e.g., ``not offensive'', ``none'', ``harmless'').
\paragraph{How good are different \modelName models?}
As shown in Table \ref{tab:model_size_results}, we observe all variants of our model have relatively high BLEU and ROUGE scores. As the model size increases, the average performance increases accordingly. It is interesting to see that \modelName (Large) achieves the best performance in the power dynamics and offensiveness dimension, which indicates that increasing modeling size does not guarantee improvement in every explanation dimension in \FrameNametext.
\paragraph{How important context is for \modelName?}
\label{sec:role-of-context}
\begin{table}[!ht]
\centering
\resizebox{0.9\linewidth}{!}{
\begin{tabular}{c  c | c  c c}
    \toprule
    \begin{tabular}[c]{@{}c@{}}Training\\ w/ context\end{tabular} & \begin{tabular}[c]{@{}c@{}}Inference\\ w/ context\end{tabular} & BLEU & ROUGE & Human* \\ \midrule
    $\times$ & $\times$ & 33.0 & 47.6 & 66.54\\
    \checkmark & $\times$ & 31.0 & 45.0 & 70.82 \\
    \checkmark & \checkmark & \textbf{35.7} & \textbf{49.5} & \textbf{75.46} \\
    \bottomrule
    \end{tabular}}
\caption{Automatic and human evaluations of context-aware and context-agnostic versions of \modelName (XL). Human evaluations are done on the same random subset (100) on all three variations.
\textbf{Takeaway:} context significantly improves \modelName both in training and inference on \datasetName. 
}
\label{tab:context-influence}
\end{table}
We examine how context influences \modelName's ability to generate explanations. 
In context-agnostic model setups, the source sequence is formatted as ``\texttt{This is a statement: [statement]}'', omitting the speaker, listener, and situation. 
As shown in Table \ref{tab:context-influence}, incorporating context at training and inference time improves \modelName's performance %
across the automatic and human evaluation. 
This is consistent with our hypothesis that context is important for understanding the toxicity of statements.

\paragraph{How well do models adapt to counterfactual contexts?}
\label{sec:exp_results_cf}

\label{sec:breakdown_analysis}
\begin{table}[!ht]
    \centering
    \resizebox{0.9\linewidth}{!}{
    \begin{tabular}{c c c c c }
    \toprule
    & Accuracy & Recall & Precision & F1 \\ \midrule
    All Toxic & 50.0 & \textbf{100.0} & 50.0  & 67.8\\
    GPT-3.5 & 55.2 & 99.4 & 52.7 & 68.9\\ \hline
    XL WoC & 50.0 & 72.3 & 50.0 & 59.1 \\
    XL & 66.5 & \textbf{98.84} & 60.0 & 74.7 \\
    XXL & \textbf{71.4} & 96.5 & \textbf{64.2} & \textbf{77.1}\\
    \bottomrule
    \end{tabular}}
    \caption{
    Accuracy, derived from binarizing the ``offensiveness'' explanation, for different models on \datasetNameCounf (WoC means Without Context).
    All Toxic means predicting every statement as toxic.
    \textbf{Takeaway:} \modelName adapts to counterfactual contexts better than GPT-3.5 (\texttt{text-davinci-003} Jan 13th 2022).
    }
    \label{tab:adv_results}
\end{table}

We then investigate how well our model, as well as GPT-3.5 ,\footnote{\texttt{text-davinci-003} Jan 13th 2022} identifies the offensiveness of statements when the context drastically alters the implications.
We then compare different models' ability to classify whether the statement is offensive or not given the counterfactual %
context in \datasetNameCounf. 

Surprisingly, although our model is only trained on the GPT-3.5-generated \datasetName, it outperforms GPT-3.5 (in a few-shot setting as described in \S\ref{sec:collect-explanations}) on \datasetNameCounf (Table \ref{tab:adv_results}). Table \ref{tab:adv_model_examples} shows some example predictions on the counterfactual context pairs. 
GPT-3.5 tends to ``over-interpret'' the statement, possibly due to the information in the prompts. For example, for the last statement in Table \ref{tab:adv_model_examples}, GPT-3.5 infers the implication as ``It implies that people of color are not typically articulate'', while such statement-context pair contains no information about people of color.
In general, counterfactual contexts are still challenging even for our best-performing models.

\section{Conclusion \& Discussion}
We introduce \FrameName frames, a formalism to distill the context-dependent implications, effects, and harms of toxic language. 
\FrameNametext captures seven explanation dimensions, inspired by frame semantics \citep{fillmore1976frame}, social bias frames \citep{sap2020socialbiasframes}, and psychology and sociolinguistics literature on social biases and prejudice \citep{nieto2006understanding,Nadal2014-ri}. 
As a step towards addressing the importance of context in content moderation, we create \datasetName, a novel dataset of toxic comments populated with contextual factors as well as explanations. 
We also build \datasetNameCounf, a small-scale, curated dataset of toxic comments paired with counterfactual contexts that significantly alter the toxicity and implication of statements.

We contribute \modelName, a new model trained with \datasetName for producing explanations of toxic statements given the statement and its social context. 
We show that modeling without contextual factors is insufficient for explaining toxicity. 
\modelName also outperforms GPT-3.5 in \datasetNameCounf, even though it is trained on data generated by GPT-3.5. 

We view \FrameName as a vital step towards addressing the importance of context in content moderation and many other social NLP tasks. 
Potential \textit{future} applications of \FrameNametext include automatic categorization of different types of offensiveness, such as hate speech, harassment, and microaggressions, as well as the development of more robust and fair content moderation systems.
Furthermore, our approach has the potential to assist content moderators by providing free-text explanations.
These explanations can help moderators better understand the rationale behind models' predictions, allowing them to make more informed decisions when reviewing flagged content \citep{zhang2023biasx}. 
This is particularly important given the growing calls for transparency and accountability in content moderation processes
\citep{hawaiiAIAssistedExplainable}. 

Besides content moderation, \FrameNametext also has the potential to test linguistic and psychological theories about offensive statements. %
While we made some preliminary attempts in this direction in \S\ref{sec:data_collection} and \S\ref{sec:adv-data-description}, 
more work is needed to fully realize this potential.
For example, future studies could investigate the differences in in-group and out-group interpretations of offensive statements, as well as the role of power dynamics, cultural background, and individual sensitivities in shaping perceptions of offensiveness.

\section*{Limitations \& Ethical and Societal Considerations}
\label{sec:limit_and_ethics}

We consider the following limitations and societal considerations of our work.

\paragraph{Machine-generated Data} Our analysis is based on GPT-3 generated data. Though not perfectly aligned with real-world scenarios, as demonstrated in \citet{Park2022SocialSC}, such analysis can provide insights into the nature of social interactions. 
However, this could induce specific biases, such as skewing towards interpretations of words aligned with GPT-3.5's training domains and potentially overlooking more specialized domains or minority speech \citep{bender21on, bommasani2021onthe}. 
The pervasive issue of bias in offensive language detection and in LLMs more generally requires exercising extra caution. 
We deliberately generate multiple contexts for every statement as an indirect means of managing the biases. 
Nevertheless, it is a compelling direction for future research to investigate the nature of biases latent in distilled contexts for harmful speech and further investigate their potential impact. 
For example, it would be valuable to collect human-annotated data on \FrameName to compare with the machine-generated data.
However, we must also recognize that humans are not immune to biases \citep{sap2019risk, sap2022annotatorsWithAttitudes}, and therefore, such investigations should be carefully designed.

\paragraph{Limited Contextual Variables} Although \datasetName has rich contexts, capturing the full context of statements is challenging. 
Future work should explore incorporating more quantitative features (e.g., the number of followers of the speaker) to supplement contextual variables such as social role and power dynamics.
In this work, we focus on the immediate context of a toxic statement. However, we recognize that the context of a toxic statement can be much longer. We have observed significant effects even in relatively brief contexts, indicating the potential for improved performance when more extended contexts are present. We believe that future research could explore the influence of richer contexts by including other modalities (e.g., images, videos, etc.).

\paragraph{Limited Identity Descriptions}
Our work focused on distilling the most salient identity characteristics that could affect the implications of toxicity of statements. 
This often resulted in generic identity labels such as ``a white person'' or ``A Black woman'' being generated without social roles.
This risks essentialism, i.e., the assumption that all members of a demographic group have inherent qualities and experiences, which can be harmful and perpetuate stereotypical thinking \cite{chen2018psychological,mandalaywala2018essentialism,kurzwelly2020allure}.
Future work should explore incorporating more specific identity descriptions that circumvent the risk of essentializing groups.

\paragraph{English Only} We only look at a US-centric perspective in our investigation.
Obviously, online hate and abuse is manifested in many languages \citep{arango-monnar-etal-2022-resources}, so we hope future work will adapt our frames to different languages and different cultures. 

\paragraph{Subjectivity in Offensiveness} Not everyone agrees that things are offensive, or has the same interpretation of offensiveness (depending on their own background and beliefs; \citealp{sap2022annotatorsWithAttitudes}). 
Our in-context prompts and qualification likely make both our machine-generated explanations and human annotations prescriptive \cite{rottger-etal-2021-hatecheck}, in contrast to a more descriptive approach where we would examine different interpretations. 
We leave that up for future work.

\paragraph{Dual Use} We aim to combat the negative effects and harms of discriminatory language on already marginalized people \citep{sap2019risk, davidson-etal-2019-racial}. 
It is possible however that our frames, dataset, and models could be used to perpetuate harm against those very people. 
We do not endorse the use of our data for those purposes.

\paragraph{Risk of Suppressing Speech} Our frames, dataset, and models are built with content moderation in mind, as online spaces are increasingly riddled with hate and abuse and content moderators are struggling to sift through all of the content. 
We hope future work will examine frameworks for using our frames to help content moderators. 
We do not endorse the use of our system to suppress speech without human oversight and encourage practitioners to take non-censorship-oriented approaches to content moderation (e.g., counterspeech \citep{tekiroglu-etal-2022-using}).

\paragraph{Harms of Exposing Workers to Toxic Content} The verification process of \datasetName and \datasetNameCounf is performed by human annotators. 
Exposure to such offensive content can be harmful to the annotators \citep{liu-etal-2016-evaluate}. 
We mitigated these by designing minimum annotation workload, paying workers above minimum wage (\$7-12), and providing them with crisis management resources. 
Our annotation work is also supervised by an Institutional Review Board (IRB).

\section*{Acknowledgements}
First of all, we thank our workers on MTurk for their hard work and thoughtful responses.
We thank the anonymous reviewers for their helpful comments. 
We also thank Shengyu Feng and members of the CMU LTI COMEDY group for their feedback, and OpenAI for providing access to the GPT-3.5 API. 
This research was supported in part by the Meta Fundamental AI Research Laboratories (FAIR) ``\textit{Dynabench Data Collection and Benchmarking Platform}'' award ``\textit{ContExTox: Context-Aware and Explainable Toxicity Detection},'' and CISCO Ethics in AI
award ``\textit{ExpHarm: Socially Aware, Ethically Informed, and Explanation-Centric AI Systems}.''

\clearpage
\bibliography{custom_clean}

\begin{thebibliography}{80}
\expandafter\ifx\csname natexlab\endcsname\relax\def\natexlab#1{#1}\fi

\bibitem[{Aguinis et~al.(2021)Aguinis, Villamor, and Ramani}]{herman21mturk}
Herman Aguinis, Isabel Villamor, and Ravi~S. Ramani. 2021.
\newblock \href {https://doi.org/10.1177/0149206320969787} {Mturk research: Review and recommendations}.
\newblock \emph{Journal of Management}, 47(4):823--837.

\bibitem[{Arango~Monnar et~al.(2022)Arango~Monnar, Perez, Poblete, Salda{\~n}a, and Proust}]{arango-monnar-etal-2022-resources}
Ayme Arango~Monnar, Jorge Perez, Barbara Poblete, Magdalena Salda{\~n}a, and Valentina Proust. 2022.
\newblock \href {https://aclanthology.org/2022.woah-1.12} {Resources for multilingual hate speech detection}.
\newblock In \emph{Proceedings of the Sixth Workshop on Online Abuse and Harms (WOAH)}.

\bibitem[{Bender and Friedman(2018)}]{bender-friedman-2018-data}
Emily~M. Bender and Batya Friedman. 2018.
\newblock \href {https://aclanthology.org/Q18-1041} {Data statements for natural language processing: Toward mitigating system bias and enabling better science}.
\newblock \emph{Transactions of the Association for Computational Linguistics}.

\bibitem[{Bender et~al.(2021)Bender, Gebru, McMillan-Major, and Shmitchell}]{bender21on}
Emily~M. Bender, Timnit Gebru, Angelina McMillan-Major, and Shmargaret Shmitchell. 2021.
\newblock \href {https://doi.org/10.1145/3442188.3445922} {On the dangers of stochastic parrots: Can language models be too big?}
\newblock In \emph{Proceedings of the 2021 ACM Conference on Fairness, Accountability, and Transparency}, FAccT '21.

\bibitem[{Bommasani et~al.(2021)Bommasani, Hudson, Adeli, Altman, Arora, von Arx, Bernstein, Bohg, Bosselut, Brunskill et~al.}]{bommasani2021onthe}
Rishi Bommasani, Drew~A. Hudson, Ehsan Adeli, Russ Altman, Simran Arora, Sydney von Arx, Michael~S. Bernstein, Jeannette Bohg, Antoine Bosselut, Emma Brunskill, et~al. 2021.
\newblock \href {https://arxiv.org/abs/2108.07258} {On the opportunities and risks of foundation models}.

\bibitem[{Bunde(2021)}]{Bunde2021-zq}
Enrico Bunde. 2021.
\newblock \href {https://scholarspace.manoa.hawaii.edu/handle/10125/70766} {{AI-Assisted} and explainable hate speech detection for social media {Moderators--A} design science approach}.
\newblock In \emph{Proceedings of the 54th Hawaii International Conference on System Sciences}.

\bibitem[{Bunde(2023)}]{hawaiiAIAssistedExplainable}
Enrico Bunde. 2023.
\newblock {A}{I}-{A}ssisted and {E}xplainable {H}ate {S}peech {D}etection for {S}ocial {M}edia {M}oderators.
\newblock \url{https://scholarspace.manoa.hawaii.edu/items/f21c8b34-5d62-40d0-919f-a4e07cfbbc32}.
\newblock [Accessed 13-May-2023].

\bibitem[{Chang et~al.(2021)Chang, He, Jia, and Singh}]{chang-etal-2021-robustness}
Kai-Wei Chang, He~He, Robin Jia, and Sameer Singh. 2021.
\newblock \href {https://aclanthology.org/2021.emnlp-tutorials.5} {Robustness and adversarial examples in natural language processing}.
\newblock In \emph{Proc. of EMNLP}.

\bibitem[{Chen and Ratliff(2018)}]{chen2018psychological}
Jacqueline~M Chen and Kate~A Ratliff. 2018.
\newblock Psychological essentialism predicts intergroup bias.
\newblock \emph{Social Cognition}, 36(3):301--323.

\bibitem[{Choi(2022)}]{Choi2022-sv}
Yejin Choi. 2022.
\newblock \href {http://dx.doi.org/10.1162/DAED_a_01906} {The curious case of commonsense intelligence}.
\newblock \emph{Daedalus}.

\bibitem[{Chung et~al.(2022)Chung, Hou, Longpre, Zoph, Tay, Fedus, Li, Wang, Dehghani, Brahma et~al.}]{chung2022scaling}
Hyung~Won Chung, Le~Hou, Shayne Longpre, Barret Zoph, Yi~Tay, William Fedus, Eric Li, Xuezhi Wang, Mostafa Dehghani, Siddhartha Brahma, et~al. 2022.
\newblock \href {https://arxiv.org/abs/2210.11416} {Scaling instruction-finetuned language models}.
\newblock \emph{ArXiv preprint}.

\bibitem[{Clark et~al.(2021)Clark, August, Serrano, Haduong, Gururangan, and Smith}]{clark-etal-2021-thats}
Elizabeth Clark, Tal August, Sofia Serrano, Nikita Haduong, Suchin Gururangan, and Noah~A. Smith. 2021.
\newblock \href {https://aclanthology.org/2021.acl-long.565} {All that{'}s {`}human{'} is not gold: Evaluating human evaluation of generated text}.
\newblock In \emph{Proc. of ACL}.

\bibitem[{Cowan and Hodge(1996)}]{cowan_judgments_1996}
Gloria Cowan and Cyndi Hodge. 1996.
\newblock \href {http://onlinelibrary.wiley.com/doi/abs/10.1111/j.1559-1816.1996.tb01854.x} {Judgments of {Hate} {Speech}: {The} {Effects} of {Target} {Group}, {Publicness}, and {Behavioral} {Responses} of the {Target}}.
\newblock \emph{Journal of Applied Social Psychology}.

\bibitem[{Cowan and Mettrick(2002)}]{cowan_effects_2002}
Gloria Cowan and Jon Mettrick. 2002.
\newblock \href {http://onlinelibrary.wiley.com/doi/abs/10.1111/j.1559-1816.2002.tb00213.x} {The effects of {Target} {Variables} and {Settting} on {Perceptions} of {Hate} {Speech1}}.
\newblock \emph{Journal of Applied Social Psychology}.

\bibitem[{Davidson et~al.(2019)Davidson, Bhattacharya, and Weber}]{davidson-etal-2019-racial}
Thomas Davidson, Debasmita Bhattacharya, and Ingmar Weber. 2019.
\newblock \href {https://aclanthology.org/W19-3504} {Racial bias in hate speech and abusive language detection datasets}.
\newblock In \emph{Proceedings of the Third Workshop on Abusive Language Online}.

\bibitem[{Davidson et~al.(2017)Davidson, Warmsley, Macy, and Weber}]{davidson_automated_2017}
Thomas Davidson, Dana Warmsley, Michael Macy, and Ingmar Weber. 2017.
\newblock \href {https://arxiv.org/pdf/1703.04009.pdf} {Automated {Hate} {Speech} {Detection} and the {Problem} of {Offensive} {Language}}.
\newblock In \emph{Proceedings of the 11th {International} {Conference} on {Web} and {Social} {Media} ({ICWSM})}.

\bibitem[{Dodge et~al.(2022)Dodge, Prewitt, Tachet~des Combes, Odmark, Schwartz, Strubell, Luccioni, Smith, DeCario, and Buchanan}]{dodge2022measuring}
Jesse Dodge, Taylor Prewitt, Remi Tachet~des Combes, Erika Odmark, Roy Schwartz, Emma Strubell, Alexandra~Sasha Luccioni, Noah~A Smith, Nicole DeCario, and Will Buchanan. 2022.
\newblock \href {https://dl.acm.org/doi/10.1145/3531146.3533234} {Measuring the carbon intensity of ai in cloud instances}.
\newblock In \emph{2022 ACM Conference on Fairness, Accountability, and Transparency}.

\bibitem[{Dynel(2015)}]{Dynel2015-kp}
Marta Dynel. 2015.
\newblock The landscape of impoliteness research.
\newblock \emph{Journal of Politeness Research}.

\bibitem[{Eliasoph and Lichterman(2003)}]{eliasoph_culture_2003}
Nina Eliasoph and Paul Lichterman. 2003.
\newblock \href {http://www.journals.uchicago.edu/doi/10.1086/367920} {Culture in {Interaction}}.
\newblock \emph{American Journal of Sociology}.

\bibitem[{ElSherief et~al.(2021)ElSherief, Ziems, Muchlinski, Anupindi, Seybolt, De~Choudhury, and Yang}]{elsherief2021latent}
Mai ElSherief, Caleb Ziems, David Muchlinski, Vaishnavi Anupindi, Jordyn Seybolt, Munmun De~Choudhury, and Diyi Yang. 2021.
\newblock \href {https://aclanthology.org/2021.emnlp-main.29} {Latent hatred: A benchmark for understanding implicit hate speech}.
\newblock In \emph{Proc. of EMNLP}.

\bibitem[{Fillmore(1976)}]{fillmore1976frame}
Charles~J. Fillmore. 1976.
\newblock \href {http://arxiv.org/abs/https://nyaspubs.onlinelibrary.wiley.com/doi/pdf/10.1111/j.1749-6632.1976.tb25467.x} {Frame semantics and the nature of language*}.
\newblock \emph{Annals of the New York Academy of Sciences}.

\bibitem[{Fine(1979)}]{fine_small_1979}
Gary~Alan Fine. 1979.
\newblock \href {http://www.jstor.org/stable/2094525?origin=crossref} {Small {Groups} and {Culture} {Creation}: {The} {Idioculture} of {Little} {League} {Baseball} {Teams}}.
\newblock \emph{American Sociological Review}.

\bibitem[{Fiske(2017)}]{fiske2017prejudices}
Susan~T Fiske. 2017.
\newblock \href {https://pubmed.ncbi.nlm.nih.gov/28972839/} {Prejudices in cultural contexts: Shared stereotypes (gender, age) versus variable stereotypes (race, ethnicity, religion)}.
\newblock \emph{Perspectives on psychological science}.

\bibitem[{Forbes et~al.(2020)Forbes, Hwang, Shwartz, Sap, and Choi}]{forbes2020socialchemistry}
Maxwell Forbes, Jena~D. Hwang, Vered Shwartz, Maarten Sap, and Yejin Choi. 2020.
\newblock \href {https://aclanthology.org/2020.emnlp-main.48} {Social chemistry 101: Learning to reason about social and moral norms}.
\newblock In \emph{Proc. of EMNLP}.

\bibitem[{Founta et~al.(2018)Founta, Djouvas, Chatzakou, Leontiadis, Blackburn, Stringhini, Vakali, Sirivianos, and Kourtellis}]{Founta2018TwitterAbusive}
Antigoni-Maria Founta, Constantinos Djouvas, Despoina Chatzakou, Ilias Leontiadis, Jeremy Blackburn, Gianluca Stringhini, Athena Vakali, Michael Sirivianos, and Nicolas Kourtellis. 2018.
\newblock \href {https://www.aaai.org/ocs/index.php/ICWSM/ICWSM18/paper/viewPaper/17909} {Large scale crowdsourcing and characterization of {Twitter} abusive behavior}.
\newblock In \emph{{ICWSM}}.

\bibitem[{Gabriel et~al.(2022)Gabriel, Hallinan, Sap, Nguyen, Roesner, Choi, and Choi}]{gabriel-etal-2022-misinfo}
Saadia Gabriel, Skyler Hallinan, Maarten Sap, Pemi Nguyen, Franziska Roesner, Eunsol Choi, and Yejin Choi. 2022.
\newblock \href {https://aclanthology.org/2022.acl-long.222} {Misinfo reaction frames: Reasoning about readers{'} reactions to news headlines}.
\newblock In \emph{Proc. of ACL}.

\bibitem[{Gardner et~al.(2020)Gardner, Artzi, Basmov, Berant, Bogin, Chen, Dasigi, Dua, Elazar, Gottumukkala, Gupta, Hajishirzi, Ilharco, Khashabi, Lin, Liu, Liu, Mulcaire, Ning, Singh, Smith, Subramanian, Tsarfaty, Wallace, Zhang, and Zhou}]{gardner-etal-2020-evaluating}
Matt Gardner, Yoav Artzi, Victoria Basmov, Jonathan Berant, Ben Bogin, Sihao Chen, Pradeep Dasigi, Dheeru Dua, Yanai Elazar, Ananth Gottumukkala, Nitish Gupta, Hannaneh Hajishirzi, Gabriel Ilharco, Daniel Khashabi, Kevin Lin, Jiangming Liu, Nelson~F. Liu, Phoebe Mulcaire, Qiang Ning, Sameer Singh, Noah~A. Smith, Sanjay Subramanian, Reut Tsarfaty, Eric Wallace, Ally Zhang, and Ben Zhou. 2020.
\newblock \href {https://aclanthology.org/2020.findings-emnlp.117} {Evaluating models{'} local decision boundaries via contrast sets}.
\newblock In \emph{Findings of the Association for Computational Linguistics: EMNLP 2020}.

\bibitem[{Gillespie et~al.(2020)Gillespie, Aufderheide, Carmi, Gerrard, Gorwa, Matamoros-Fernandez, Roberts, Sinnreich, and West}]{Gillespie2020-aw}
Tarleton Gillespie, Patricia Aufderheide, Elinor Carmi, Ysabel Gerrard, Robert Gorwa, Ariadna Matamoros-Fernandez, Sarah~T Roberts, Aram Sinnreich, and Sarah~Myers West. 2020.
\newblock \href {https://eprints.qut.edu.au/205685/} {Expanding the debate about content moderation: Scholarly research agendas for the coming policy debates}.
\newblock \emph{Internet Policy Review}.

\bibitem[{Goodman and Frank(2016)}]{goodman2016pragmatic}
Noah~D Goodman and Michael~C Frank. 2016.
\newblock \href {https://www.sciencedirect.com/science/article/pii/S136466131630122X} {Pragmatic language interpretation as probabilistic inference}.
\newblock \emph{Trends in cognitive sciences}.

\bibitem[{Gregory(1967)}]{gregory_1967}
Michael Gregory. 1967.
\newblock \href {https://www.jstor.org/stable/4174964} {Aspects of varieties differentiation}.
\newblock \emph{Journal of Linguistics}.

\bibitem[{Grice(1975)}]{grice1975logic}
Herbert~P Grice. 1975.
\newblock \href {https://www.ucl.ac.uk/ls/studypacks/Grice-Logic.pdf} {Logic and conversation}.
\newblock In \emph{Speech acts}. Brill.

\bibitem[{Hartvigsen et~al.(2022)Hartvigsen, Gabriel, Palangi, Sap, Ray, and Kamar}]{hartvigsen2022toxigen}
Thomas Hartvigsen, Saadia Gabriel, Hamid Palangi, Maarten Sap, Dipankar Ray, and Ece Kamar. 2022.
\newblock \href {https://aclanthology.org/2022.acl-long.234.pdf} {Toxigen: Controlling language models to generate implied and adversarial toxicity}.
\newblock In \emph{ACL}.

\bibitem[{Hashimoto et~al.(2019)Hashimoto, Zhang, and Liang}]{hashimoto-etal-2019-unifying}
Tatsunori~B. Hashimoto, Hugh Zhang, and Percy Liang. 2019.
\newblock \href {https://aclanthology.org/N19-1169} {Unifying human and statistical evaluation for natural language generation}.
\newblock In \emph{Proc. of NAACL-HLT}.

\bibitem[{Holgate et~al.(2018)Holgate, Cachola, Preo{\c{t}}iuc-Pietro, and Li}]{holgate_why_2018}
Eric Holgate, Isabel Cachola, Daniel Preo{\c{t}}iuc-Pietro, and Junyi~Jessy Li. 2018.
\newblock \href {https://aclanthology.org/D18-1471} {Why swear? analyzing and inferring the intentions of vulgar expressions}.
\newblock In \emph{Proc. of EMNLP}.

\bibitem[{Hovy and Yang(2021)}]{hovy2021importance}
Dirk Hovy and Diyi Yang. 2021.
\newblock \href {https://aclanthology.org/2021.naacl-main.49} {The importance of modeling social factors of language: Theory and practice}.
\newblock In \emph{Proceedings of the 2021 Conference of the North American Chapter of the Association for Computational Linguistics: Human Language Technologies}.

\bibitem[{Jay and Janschewitz(2008)}]{Jay2008-xj}
Timothy Jay and Kristin Janschewitz. 2008.
\newblock \href {https://www.mcla.edu/Assets/MCLA-Files/Academics/Undergraduate/Psychology/Pragmaticsofswearing.pdf} {The pragmatics of swearing}.
\newblock \emph{Journal of Politeness Research}.

\bibitem[{Jiang et~al.(2021)Jiang, Hwang, Bhagavatula, Bras, Liang, Dodge, Sakaguchi, Forbes, Borchardt, Gabriel, Tsvetkov, Etzioni, Sap, Rini, and Choi}]{Jiang2021-tt}
Liwei Jiang, Jena~D. Hwang, Chandra Bhagavatula, Ronan~Le Bras, Jenny Liang, Jesse Dodge, Keisuke Sakaguchi, Maxwell Forbes, Jon Borchardt, Saadia Gabriel, Yulia Tsvetkov, Oren Etzioni, Maarten Sap, Regina Rini, and Yejin Choi. 2021.
\newblock \href {https://arxiv.org/abs/2110.07574} {Can machines learn morality? the delphi experiment}.

\bibitem[{Jurgens et~al.(2019)Jurgens, Hemphill, and Chandrasekharan}]{jurgens-etal-2019-just}
David Jurgens, Libby Hemphill, and Eshwar Chandrasekharan. 2019.
\newblock \href {https://aclanthology.org/P19-1357} {A just and comprehensive strategy for using {NLP} to address online abuse}.
\newblock In \emph{Proc. of ACL}.

\bibitem[{Kasper(1990)}]{KASPER1990193}
Gabriele Kasper. 1990.
\newblock \href {https://www.sciencedirect.com/science/article/pii/037821669090080W} {Linguistic politeness:: Current research issues}.
\newblock \emph{Journal of Pragmatics}.
\newblock Special Issue on Politeness.

\bibitem[{Keller and Leerssen(2020)}]{Keller2019-bo}
Daphne Keller and Paddy Leerssen. 2020.
\newblock \href {https://www.cambridge.org/core/books/social-media-and-democracy/facts-and-where-to-find-them-empirical-research-on-internet-platforms-and-content-moderation/78DE9202F2D00F2967EFC5CBDCE2CAF0} {Facts and where to find them: Empirical research on internet platforms and content moderation}.
\newblock In \emph{Social Media and Democracy}. Cambridge University Press.

\bibitem[{Khurana et~al.(2022)Khurana, Vermeulen, Nalisnick, Van~Noorloos, and Fokkens}]{khurana-etal-2022-hate}
Urja Khurana, Ivar Vermeulen, Eric Nalisnick, Marloes Van~Noorloos, and Antske Fokkens. 2022.
\newblock \href {https://aclanthology.org/2022.woah-1.17} {Hate speech criteria: A modular approach to task-specific hate speech definitions}.
\newblock In \emph{Proceedings of the Sixth Workshop on Online Abuse and Harms (WOAH)}.

\bibitem[{Kim et~al.(2022{\natexlab{a}})Kim, Hessel, Jiang, Lu, Yu, Zhou, Bras, Alikhani, Kim, Sap, and Choi}]{kim2022soda}
Hyunwoo Kim, Jack Hessel, Liwei Jiang, Ximing Lu, Youngjae Yu, Pei Zhou, Ronan~Le Bras, Malihe Alikhani, Gunhee Kim, Maarten Sap, and Yejin Choi. 2022{\natexlab{a}}.
\newblock \href {https://arxiv.org/abs/2212.10465} {Soda: Million-scale dialogue distillation with social commonsense contextualization}.
\newblock \emph{ArXiv preprint}.

\bibitem[{Kim et~al.(2022{\natexlab{b}})Kim, Yu, Jiang, Lu, Khashabi, Kim, Choi, and Sap}]{kim2022prosocialDialog}
Hyunwoo Kim, Youngjae Yu, Liwei Jiang, Ximing Lu, Daniel Khashabi, Gunhee Kim, Yejin Choi, and Maarten Sap. 2022{\natexlab{b}}.
\newblock \href {https://arxiv.org/abs/2205.12688} {Prosocialdialog: A prosocial backbone for conversational agents}.
\newblock \emph{ArXiv preprint}.

\bibitem[{Kohli et~al.(2018)Kohli, Arteaga, and McGovern}]{Kohli2018-de}
Rita Kohli, Nallely Arteaga, and Elexia~R McGovern. 2018.
\newblock \href {https://onlinelibrary.wiley.com/doi/abs/10.1002/9781119466642.ch17} {``compliments'' and ``jokes'': Unpacking racial microaggressions in the {K-12} classroom}.
\newblock In \emph{Microaggression Theory Influence and Implications}. John Wiley \& Sons.

\bibitem[{Kurzwelly et~al.(2020)Kurzwelly, Fernana, and Ngum}]{kurzwelly2020allure}
Jonatan Kurzwelly, Hamid Fernana, and Muhammad~Elvis Ngum. 2020.
\newblock The allure of essentialism and extremist ideologies.
\newblock \emph{Anthropology Southern Africa}, 43(2):107--118.

\bibitem[{Lai et~al.(2022)Lai, Carton, Bhatnagar, Liao, Zhang, and Tan}]{Lai2022-ns}
Vivian Lai, Samuel Carton, Rajat Bhatnagar, Q~Vera Liao, Yunfeng Zhang, and Chenhao Tan. 2022.
\newblock \href {https://dl.acm.org/doi/10.1145/3491102.3501999} {{Human-AI} collaboration via conditional delegation: A case study of content moderation}.
\newblock In \emph{{CHI}}.

\bibitem[{Li et~al.(2020)Li, Shengshuo, Liu, Wu, Zhou, and Steinert-Threlkeld}]{li-etal-2020-linguistically}
Chuanrong Li, Lin Shengshuo, Zeyu Liu, Xinyi Wu, Xuhui Zhou, and Shane Steinert-Threlkeld. 2020.
\newblock \href {https://aclanthology.org/2020.blackboxnlp-1.12} {Linguistically-informed transformations ({LIT}): A method for automatically generating contrast sets}.
\newblock In \emph{Proceedings of the Third BlackboxNLP Workshop on Analyzing and Interpreting Neural Networks for NLP}.

\bibitem[{Liu et~al.(2022)Liu, Swayamdipta, Smith, and Choi}]{liu2022wanli}
Alisa Liu, Swabha Swayamdipta, Noah~A. Smith, and Yejin Choi. 2022.
\newblock \href {https://aclanthology.org/2022.findings-emnlp.508} {{WANLI}: Worker and {AI} collaboration for natural language inference dataset creation}.
\newblock In \emph{In proc. of Findings of EMNLP}. Association for Computational Linguistics.

\bibitem[{Liu et~al.(2016)Liu, Lowe, Serban, Noseworthy, Charlin, and Pineau}]{liu-etal-2016-evaluate}
Chia-Wei Liu, Ryan Lowe, Iulian Serban, Mike Noseworthy, Laurent Charlin, and Joelle Pineau. 2016.
\newblock \href {https://aclanthology.org/D16-1230} {How {NOT} to evaluate your dialogue system: An empirical study of unsupervised evaluation metrics for dialogue response generation}.
\newblock In \emph{Proc. of EMNLP}.

\bibitem[{Mandalaywala et~al.(2018)Mandalaywala, Amodio, and Rhodes}]{mandalaywala2018essentialism}
Tara~M Mandalaywala, David~M Amodio, and Marjorie Rhodes. 2018.
\newblock Essentialism promotes racial prejudice by increasing endorsement of social hierarchies.
\newblock \emph{Social Psychological and Personality Science}, 9(4):461--469.

\bibitem[{Miller(2019)}]{Miller2019-fm}
Tim Miller. 2019.
\newblock \href {https://www.sciencedirect.com/science/article/pii/S0004370218305988} {Explanation in artificial intelligence: Insights from the social sciences}.
\newblock \emph{Artificial intelligence}.

\bibitem[{Monroe and Potts(2015)}]{will2015learning}
Will Monroe and Christopher Potts. 2015.
\newblock \href {https://arxiv.org/abs/1510.06807} {Learning in the rational speech acts model}.
\newblock \emph{ArXiv preprint}.

\bibitem[{Nadal et~al.(2014)Nadal, Davidoff, Davis, and Wong}]{Nadal2014-ri}
Kevin~L Nadal, Kristin~C Davidoff, Lindsey~S Davis, and Yinglee Wong. 2014.
\newblock Emotional, behavioral, and cognitive reactions to microaggressions: Transgender perspectives.
\newblock \emph{Psychology of Sexual Orientation and Gender Diversity}.

\bibitem[{Nieto and Boyer(2006)}]{nieto2006understanding}
Leticia Nieto and Margot Boyer. 2006.
\newblock \href {https://beyondinclusionbeyondempowerment.com/wp-content/uploads/2019/12/nieto-articles-understanding-oppression-2006.pdf} {Understanding oppression: Strategies in addressing power and privilege}.
\newblock \emph{Colors NW}.

\bibitem[{Ouyang et~al.(2022)Ouyang, Wu, Jiang, Almeida, Wainwright, Mishkin, Zhang, Agarwal, Slama, Ray, Schulman, Hilton, Kelton, Miller, Simens, Askell, Welinder, Christiano, Leike, and Lowe}]{Ouyang2021InstructGPT}
Long Ouyang, Jeff Wu, Xu~Jiang, Diogo Almeida, Carroll~L. Wainwright, Pamela Mishkin, Chong Zhang, Sandhini Agarwal, Katarina Slama, Alex Ray, John Schulman, Jacob Hilton, Fraser Kelton, Luke Miller, Maddie Simens, Amanda Askell, Peter Welinder, Paul Christiano, Jan Leike, and Ryan Lowe. 2022.
\newblock \href {https://arxiv.org/abs/2203.02155} {Training language models to follow instructions with human feedback}.

\bibitem[{Park et~al.(2022)Park, Popowski, Cai, Morris, Liang, and Bernstein}]{Park2022SocialSC}
Joon~Sung Park, Lindsay Popowski, Carrie~J. Cai, Meredith~Ringel Morris, Percy Liang, and Michael~S. Bernstein. 2022.
\newblock \href {https://dl.acm.org/doi/abs/10.1145/3526113.3545616} {Social simulacra: Creating populated prototypes for social computing systems}.
\newblock \emph{Proceedings of the 35th Annual ACM Symposium on User Interface Software and Technology}.

\bibitem[{Perez~Gomez(2020)}]{Perez_Gomez2020-im}
Javiera Perez~Gomez. 2020.
\newblock \href {http://dx.doi.org/10.1111/jopp.12243} {Verbal microaggressions as hyper‐implicatures}.
\newblock \emph{J. Polit. Philos.}

\bibitem[{Randolph(2005)}]{Randolph2005-iw}
Justus~J Randolph. 2005.
\newblock \href {https://dl.acm.org/doi/abs/10.1145/3526113.3545616} {{Free-Marginal} multirater kappa (multirater k[free]): An alternative to fleiss' {Fixed-Marginal} multirater kappa}.
\newblock In \emph{Proceedings of JLIS}.

\bibitem[{Ribeiro et~al.(2017)Ribeiro, Calais, Santos, Almeida, and Meira~Jr}]{ribeiro_characterizing_2017}
Manoel~Horta Ribeiro, Pedro~H. Calais, Yuri~A. Santos, Virgílio A.~F. Almeida, and Wagner Meira~Jr. 2017.
\newblock \href {https://doi.org/10.1609/icwsm.v12i1.15057} {Characterizing and {Detecting} {Hateful} {Users} on {Twitter}}.
\newblock In \emph{Proceedings of the {International} {AAAI} {Conference} on {Web} and {Social} {Media}}.
\newblock ArXiv: 1801.00317.

\bibitem[{R{\"o}ttger et~al.(2021)R{\"o}ttger, Vidgen, Nguyen, Waseem, Margetts, and Pierrehumbert}]{rottger-etal-2021-hatecheck}
Paul R{\"o}ttger, Bertie Vidgen, Dong Nguyen, Zeerak Waseem, Helen Margetts, and Janet Pierrehumbert. 2021.
\newblock \href {https://aclanthology.org/2021.acl-long.4} {{H}ate{C}heck: Functional tests for hate speech detection models}.
\newblock In \emph{Proc. of ACL}.

\bibitem[{Sap et~al.(2019{\natexlab{a}})Sap, Bras, Allaway, Bhagavatula, Lourie, Rashkin, Roof, Smith, and Choi}]{sap2019atomic}
Maarten Sap, Ronan~Le Bras, Emily Allaway, Chandra Bhagavatula, Nicholas Lourie, Hannah Rashkin, Brendan Roof, Noah~A. Smith, and Yejin Choi. 2019{\natexlab{a}}.
\newblock \href {https://doi.org/10.1609/aaai.v33i01.33013027} {{ATOMIC:} an atlas of machine commonsense for if-then reasoning}.
\newblock In \emph{The Thirty-Third {AAAI} Conference on Artificial Intelligence, {AAAI} 2019, The Thirty-First Innovative Applications of Artificial Intelligence Conference, {IAAI} 2019, The Ninth {AAAI} Symposium on Educational Advances in Artificial Intelligence, {EAAI} 2019, Honolulu, Hawaii, USA, January 27 - February 1, 2019}.

\bibitem[{Sap et~al.(2019{\natexlab{b}})Sap, Card, Gabriel, Choi, and Smith}]{sap2019risk}
Maarten Sap, Dallas Card, Saadia Gabriel, Yejin Choi, and Noah~A. Smith. 2019{\natexlab{b}}.
\newblock \href {https://aclanthology.org/P19-1163} {The risk of racial bias in hate speech detection}.
\newblock In \emph{Proc. of ACL}.

\bibitem[{Sap et~al.(2020)Sap, Gabriel, Qin, Jurafsky, Smith, and Choi}]{sap2020socialbiasframes}
Maarten Sap, Saadia Gabriel, Lianhui Qin, Dan Jurafsky, Noah~A. Smith, and Yejin Choi. 2020.
\newblock \href {https://aclanthology.org/2020.acl-main.486} {Social bias frames: Reasoning about social and power implications of language}.
\newblock In \emph{Proc. of ACL}.

\bibitem[{Sap et~al.(2022)Sap, Swayamdipta, Vianna, Zhou, Choi, and Smith}]{sap2022annotatorsWithAttitudes}
Maarten Sap, Swabha Swayamdipta, Laura Vianna, Xuhui Zhou, Yejin Choi, and Noah~A. Smith. 2022.
\newblock \href {https://aclanthology.org/2022.naacl-main.431} {Annotators with attitudes: How annotator beliefs and identities bias toxic language detection}.
\newblock In \emph{Proceedings of the 2022 Conference of the North American Chapter of the Association for Computational Linguistics: Human Language Technologies}.

\bibitem[{Strubell et~al.(2019)Strubell, Ganesh, and McCallum}]{strubell2019energy}
Emma Strubell, Ananya Ganesh, and Andrew McCallum. 2019.
\newblock \href {https://aclanthology.org/P19-1355} {Energy and policy considerations for deep learning in {NLP}}.
\newblock In \emph{Proc. of ACL}.

\bibitem[{Sue(2010)}]{sue2010microaggressions}
Derald~Wing Sue. 2010.
\newblock \href {https://www.wiley.com/en-us/Microaggressions+in+Everyday+Life:+Race,+Gender,+and+Sexual+Orientation-p-9780470491409} {\emph{Microaggressions in everyday life: Race, gender, and sexual orientation}}.
\newblock John Wiley \& Sons.

\bibitem[{Taddeo et~al.(2021)Taddeo, Tsamados, Cowls, and Floridi}]{taddeo2021artificial}
Mariarosaria Taddeo, Andreas Tsamados, Josh Cowls, and Luciano Floridi. 2021.
\newblock \href {https://papers.ssrn.com/sol3/papers.cfm?abstract_id=3873881} {Artificial intelligence and the climate emergency: Opportunities, challenges, and recommendations}.
\newblock \emph{One Earth}.

\bibitem[{Tatman(2020)}]{Tatman2020-ze}
Rachael Tatman. 2020.
\newblock \href {https://www.rctatman.com/talks/what-i-wont-build} {What {I} won't build}.
\newblock Widening NLP Workshop.

\bibitem[{Tekiro{\u{g}}lu et~al.(2022)Tekiro{\u{g}}lu, Bonaldi, Fanton, and Guerini}]{tekiroglu-etal-2022-using}
Serra~Sinem Tekiro{\u{g}}lu, Helena Bonaldi, Margherita Fanton, and Marco Guerini. 2022.
\newblock \href {https://aclanthology.org/2022.findings-acl.245} {Using pre-trained language models for producing counter narratives against hate speech: a comparative study}.
\newblock In \emph{Findings of the Association for Computational Linguistics: ACL 2022}.

\bibitem[{Vaccaro et~al.(2020)Vaccaro, Sandvig, and Karahalios}]{Vaccaro2020at}
Kristen Vaccaro, Christian Sandvig, and Karrie Karahalios. 2020.
\newblock \href {https://doi.org/10.1145/3415238} {"at the end of the day facebook does what itwants": How users experience contesting algorithmic content moderation}.
\newblock \emph{Proc. ACM Hum.-Comput. Interact.}

\bibitem[{Vidgen et~al.(2021{\natexlab{a}})Vidgen, Nguyen, Margetts, Rossini, and Tromble}]{vidgen_introducing_2021}
Bertie Vidgen, Dong Nguyen, Helen Margetts, Patricia Rossini, and Rebekah Tromble. 2021{\natexlab{a}}.
\newblock \href {https://aclanthology.org/2021.naacl-main.182} {Introducing {CAD}: the contextual abuse dataset}.
\newblock In \emph{Proceedings of the 2021 Conference of the North American Chapter of the Association for Computational Linguistics: Human Language Technologies}.

\bibitem[{Vidgen et~al.(2021{\natexlab{b}})Vidgen, Thrush, Waseem, and Kiela}]{vidgen-etal-2021-learning}
Bertie Vidgen, Tristan Thrush, Zeerak Waseem, and Douwe Kiela. 2021{\natexlab{b}}.
\newblock \href {https://aclanthology.org/2021.acl-long.132} {Learning from the worst: Dynamically generated datasets to improve online hate detection}.
\newblock In \emph{Proc. of ACL}.

\bibitem[{Waseem et~al.(2017)Waseem, Davidson, Warmsley, and Weber}]{waseem_understanding_2017}
Zeerak Waseem, Thomas Davidson, Dana Warmsley, and Ingmar Weber. 2017.
\newblock \href {https://aclanthology.org/W17-3012} {Understanding abuse: A typology of abusive language detection subtasks}.
\newblock In \emph{Proceedings of the First Workshop on Abusive Language Online}.

\bibitem[{West et~al.(2022)West, Bhagavatula, Hessel, Hwang, Jiang, Le~Bras, Lu, Welleck, and Choi}]{west-etal-2022-symbolic}
Peter West, Chandra Bhagavatula, Jack Hessel, Jena Hwang, Liwei Jiang, Ronan Le~Bras, Ximing Lu, Sean Welleck, and Yejin Choi. 2022.
\newblock \href {https://aclanthology.org/2022.naacl-main.341} {Symbolic knowledge distillation: from general language models to commonsense models}.
\newblock In \emph{Proceedings of the 2022 Conference of the North American Chapter of the Association for Computational Linguistics: Human Language Technologies}.

\bibitem[{Wiegreffe et~al.(2021)Wiegreffe, Marasovi{\'c}, and Smith}]{wiegreffe2021measuring}
Sarah Wiegreffe, Ana Marasovi{\'c}, and Noah~A. Smith. 2021.
\newblock \href {https://aclanthology.org/2021.emnlp-main.804} {{M}easuring association between labels and free-text rationales}.
\newblock In \emph{Proc. of EMNLP}.

\bibitem[{Zampieri et~al.(2019)Zampieri, Malmasi, Nakov, Rosenthal, Farra, and Kumar}]{zampieri-etal-2019-predicting}
Marcos Zampieri, Shervin Malmasi, Preslav Nakov, Sara Rosenthal, Noura Farra, and Ritesh Kumar. 2019.
\newblock \href {https://aclanthology.org/N19-1144} {Predicting the type and target of offensive posts in social media}.
\newblock In \emph{Proc. of NAACL-HLT}.

\bibitem[{Zhang et~al.(2020)Zhang, Kishore, Wu, Weinberger, and Artzi}]{Zhang2020BERTScore}
Tianyi Zhang, Varsha Kishore, Felix Wu, Kilian~Q. Weinberger, and Yoav Artzi. 2020.
\newblock \href {https://openreview.net/forum?id=SkeHuCVFDr} {Bertscore: Evaluating text generation with {BERT}}.
\newblock In \emph{Proc. of ICLR}.

\bibitem[{Zhang et~al.(2023)Zhang, Nanduri, Jiang, Wu, and Sap}]{zhang2023biasx}
Yiming Zhang, Sravani~U. Nanduri, Liwei Jiang, Tongshuang Wu, and Maarten Sap. 2023.
\newblock \href {https://arxiv.org/abs/2305.13589} {"thinking slow” in toxic language annotation with explanations of implied social biases}.
\newblock \emph{arXiv}.

\bibitem[{Zhou et~al.(2022)Zhou, Deng, Mi, Li, Wang, Huang, Jiang, Liu, and Meng}]{Zhou2022-lv}
Jingyan Zhou, Jiawen Deng, Fei Mi, Yitong Li, Yasheng Wang, Minlie Huang, Xin Jiang, Qun Liu, and Helen Meng. 2022.
\newblock \href {https://arxiv.org/abs/2202.08011} {Towards identifying social bias in dialog systems: Frame, datasets, and benchmarks}.

\bibitem[{Zhou et~al.(2021)Zhou, Sap, Swayamdipta, Choi, and Smith}]{zhou2021challenges}
Xuhui Zhou, Maarten Sap, Swabha Swayamdipta, Yejin Choi, and Noah Smith. 2021.
\newblock \href {https://aclanthology.org/2021.eacl-main.274} {Challenges in automated debiasing for toxic language detection}.
\newblock In \emph{Proceedings of the 16th Conference of the European Chapter of the Association for Computational Linguistics: Main Volume}.

\end{thebibliography}
\bibliographystyle{acl_natbib}

\appendix

\clearpage
\onecolumn
\section{Crowd-sourcing on MTurk}
\label{appendix:crowd-sourcing}
In this paper, human annotation is widely used in \S\ref{sec: collect plausible contexts}, \S\ref{sec:collect-explanations}, \S\ref{sec:adv-data-description}, \S\ref{sec:context-increase-toxic}, \S\ref{sec:role-of-context}, and \S\ref{sec:breakdown_analysis}. We restrict our worker candidates' location to U.S. and Canada and ask the workers to optionally provide coarse-grained demographic information. Among 300 candidates, 109 workers pass the qualification tests. Note that we not only give the workers scores based on their accuracy in our tests, but also manually verify their provided suggestions for explanations. Annotators are compensated \$12.8 per hour on average. The data collection procedure was approved by our institution's IRB. 

\subsection{Annotator demographics}
Due to the subjective nature of toxic language \citep{sap2022annotatorsWithAttitudes}, we aim to collect a diverse set of annotators. In our final pool of 109 annotators, the average age is 36 (ranging from 18 to 65). For political orientation, we have 64/21/24 annotators identified as liberal/conservative/neutral, respectively.
For gender identity, we have 61/46/2 annotators identify as man/woman/non-binary, respectively. There are also 40 annotators that self-identified as being part of a minority group.

\subsection{Annotation interface and instructions}
\label{appendix: Annotation interface}
\begin{figure*}
\centering
\includegraphics[width=\textwidth]{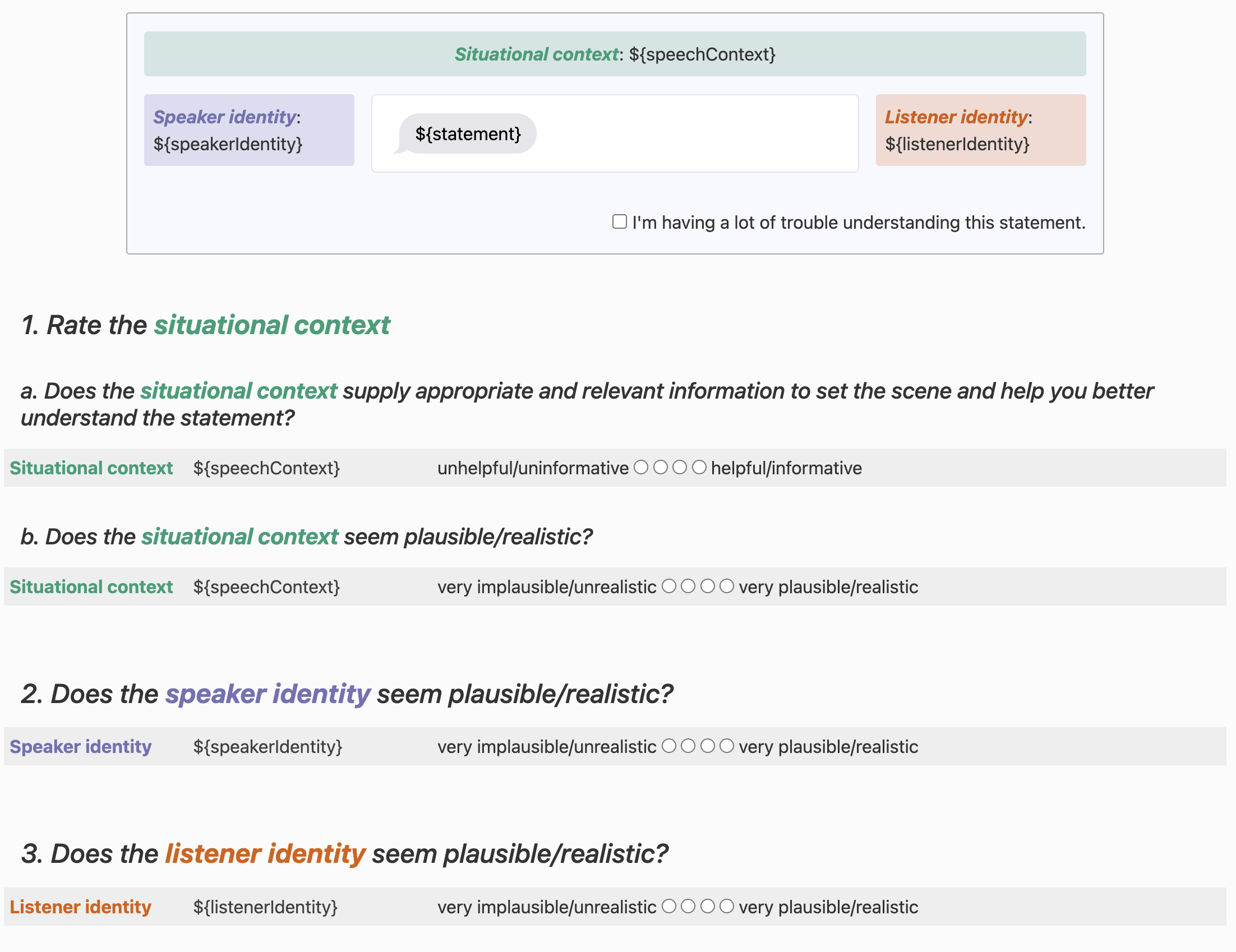}
\caption{The annotation snippet of collecting plausible scenarios (\S\ref{sec: collect plausible contexts})}
\label{fig:scenario}
\end{figure*}

\begin{figure*}
\centering
\includegraphics[width=\textwidth]{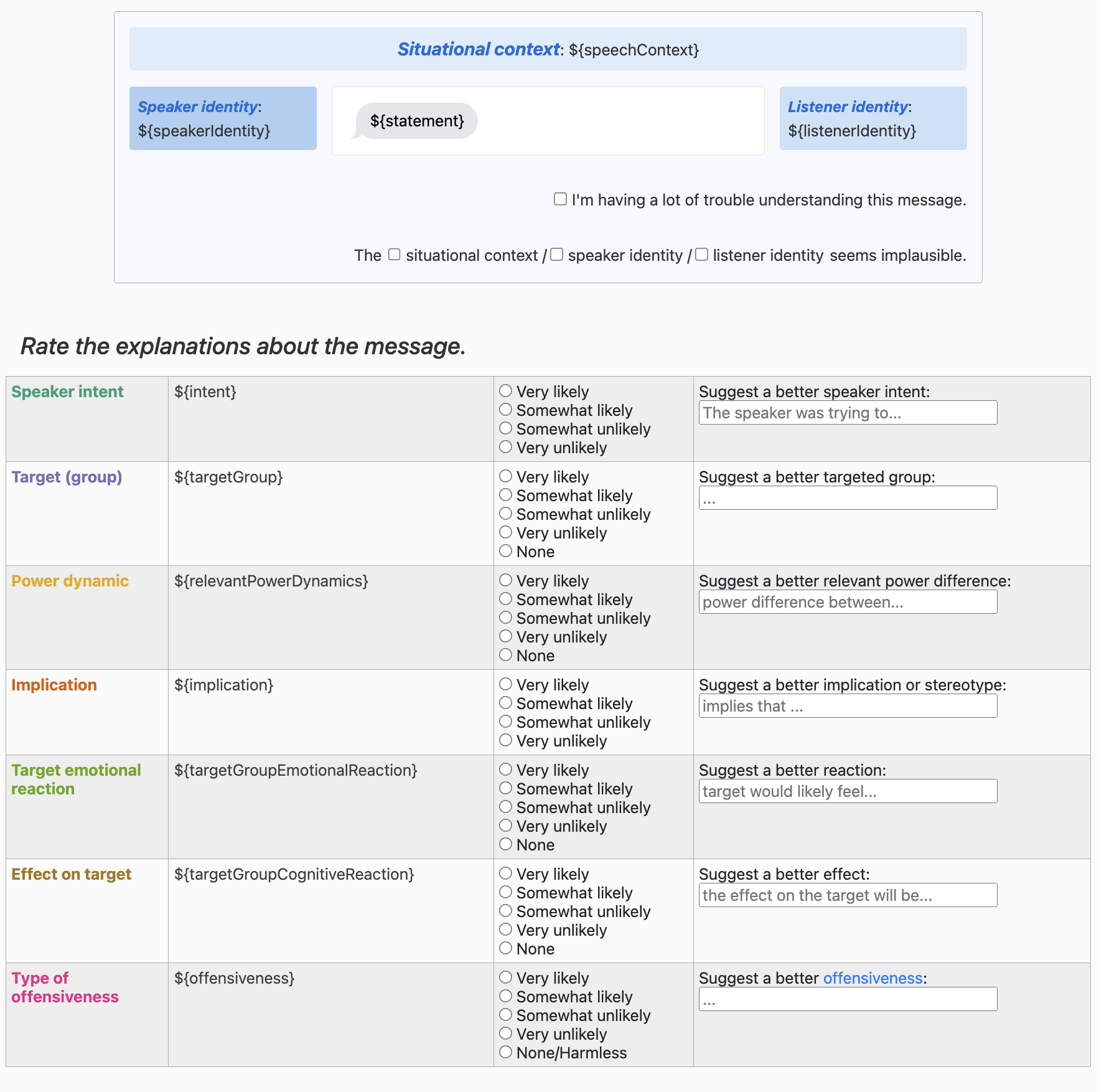}
\caption{The annotation snippet of collecting explanations (\S\ref{sec: collect plausible contexts})}
\label{fig:explanations}
\end{figure*}

\begin{figure*}
\centering
\includegraphics[width=\textwidth]{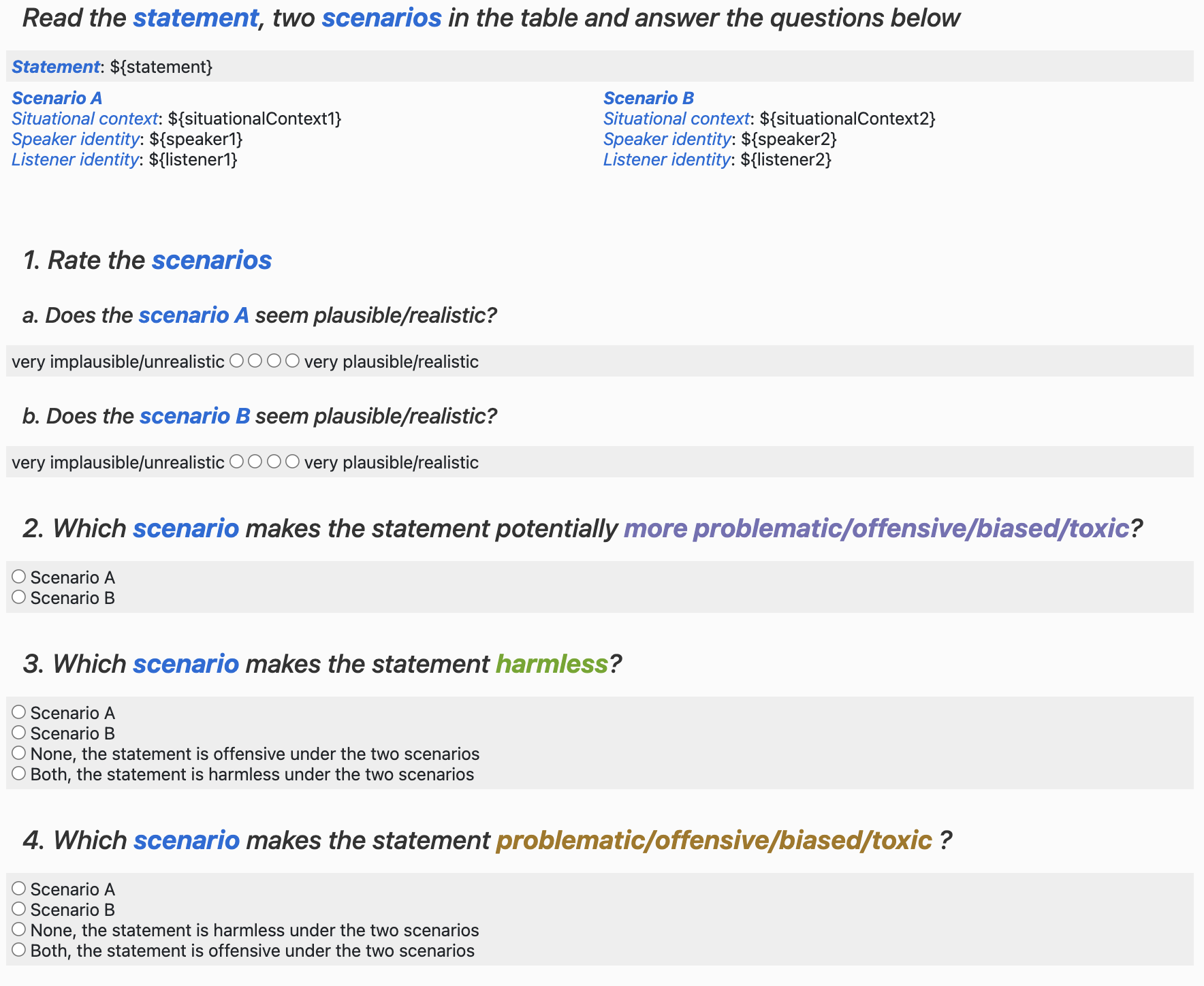}
\caption{The annotation snippet of collecting adversarial examples (\S\ref{sec:adv-data-description})}
\label{fig:adv_annotate}
\end{figure*}

As recommended by \citep{herman21mturk}, we design the MTurk interface with clear instructions, examples with explanations. The annotation snippet of collecting plausible scenarios (\S\ref{sec: collect plausible contexts}) is in Figure \ref{fig:scenario}. The annotation snippet of collecting explanations (\S\ref{sec:collect-explanations}) is in Figure \ref{fig:explanations}. The annotation snippet of collecting adversarial examples (\S\ref{sec:adv-data-description}) is in Figure \ref{fig:adv_annotate}.

\section{\modelName experiment details}

\subsection{Training details}
\label{appendix:training_details}
With the HuggingFace’s Transformers library\footnote{https://github.com/huggingface/transformers}, different variants of FLAN-T5, small, base, large, XL and XXL, are finetuned on the \FrameNametext training set for two epochs with AdamW optimizer with a learning rate of $1e^{-4}$ and batch size of 16. We use beam search as the decoding algorithm and all reported results are based on a single run. We also train a XL model using the same architecture and hyperparameters but without the context information. The sizes of \modelName range from 80M to 11B, the largest of which takes 10 hours to train in FP32 on 5 A6000 GPUs with NVLink, and can do inference in FP16 on a single A6000 GPU. We used HuggingFace evaluate package to evaluate the BLEU-2 and ROUGE-L scores. 

\subsection{Evaluation details}
\label{appendix:evaluation_details}

See Table \ref{tab:model_size_results_bertscore} for the BERTScore metrics across different model sizes.

\begin{table*}[t]
    \centering
    \resizebox*{0.95\textwidth}{!}{
        \begin{tabular}{@{}
            >{\columncolor[HTML]{FFFFFF}}r |
            >{\columncolor[HTML]{d9eee8}}c 
            >{\columncolor[HTML]{f9e4d5}}c 
            >{\columncolor[HTML]{EDE0FE}}c  
            >{\columncolor[HTML]{fbdbec}}c  
            >{\columncolor[HTML]{fbf1d5}}c 
            >{\columncolor[HTML]{e5f0d9}}c
            >{\columncolor[HTML]{f0e8d9}}c |
            >{\columncolor[HTML]{EFEFEF}}r}
            \toprule
            \multicolumn{1}{c|}{\cellcolor[HTML]{FFFFFF}} & \multicolumn{1}{c}{\cellcolor[HTML]{d9eee8}Intent} & \multicolumn{1}{c}{\cellcolor[HTML]{f9e4d5}Target group} & \multicolumn{1}{c}{\cellcolor[HTML]{EDE0FE}Power Dynamics} & \multicolumn{1}{c}{\cellcolor[HTML]{fbdbec}Implication} & \multicolumn{1}{c}{\cellcolor[HTML]{fbf1d5}Emotional React.} & \multicolumn{1}{c}{\cellcolor[HTML]{e5f0d9}Cognitive React.} & \multicolumn{1}{c|}{\cellcolor[HTML]{f0e8d9}Offensiveness} & \multicolumn{1}{c}{\cellcolor[HTML]{EFEFEF}Average} \\
            \midrule
            Small & 0.936 & 0.929 & 0.932 & 0.900 & 0.886 & 0.877 & 0.889 & 0.907 \\
            Base & 0.939 & 0.933 & 0.932 & 0.907 & 0.892 & 0.880 & 0.890 & 0.910 \\
            Large & 0.944 & 0.939 & \textbf{0.938} & 0.916 & \textbf{0.898} & \textbf{0.887} & 0.897 & 0.917 \\
            XL & 0.947 & \textbf{0.940} & \textbf{0.938} & 0.917 & 0.897 & 0.886 & \textbf{0.899} & \textbf{0.918} \\
            XXL & \textbf{0.948} & 0.939 & 0.937 & \textbf{0.918} & \textbf{0.898} & \textbf{0.887} & 0.895 & 0.917 \\
            \bottomrule
        \end{tabular}
    }
    \caption{BERTScore of different model sizes measured with automatic evaluation metrics, broken down by explanation dimension.
    }
    \label{tab:model_size_results_bertscore}
    \end{table*}

\section{GPT-3 prompts used in this paper}
\label{appendix:gpt3_prompts}
\begin{figure*}
\centering
\includegraphics[width=\textwidth, trim={4cm 3cm 4cm 3cm}, clip]{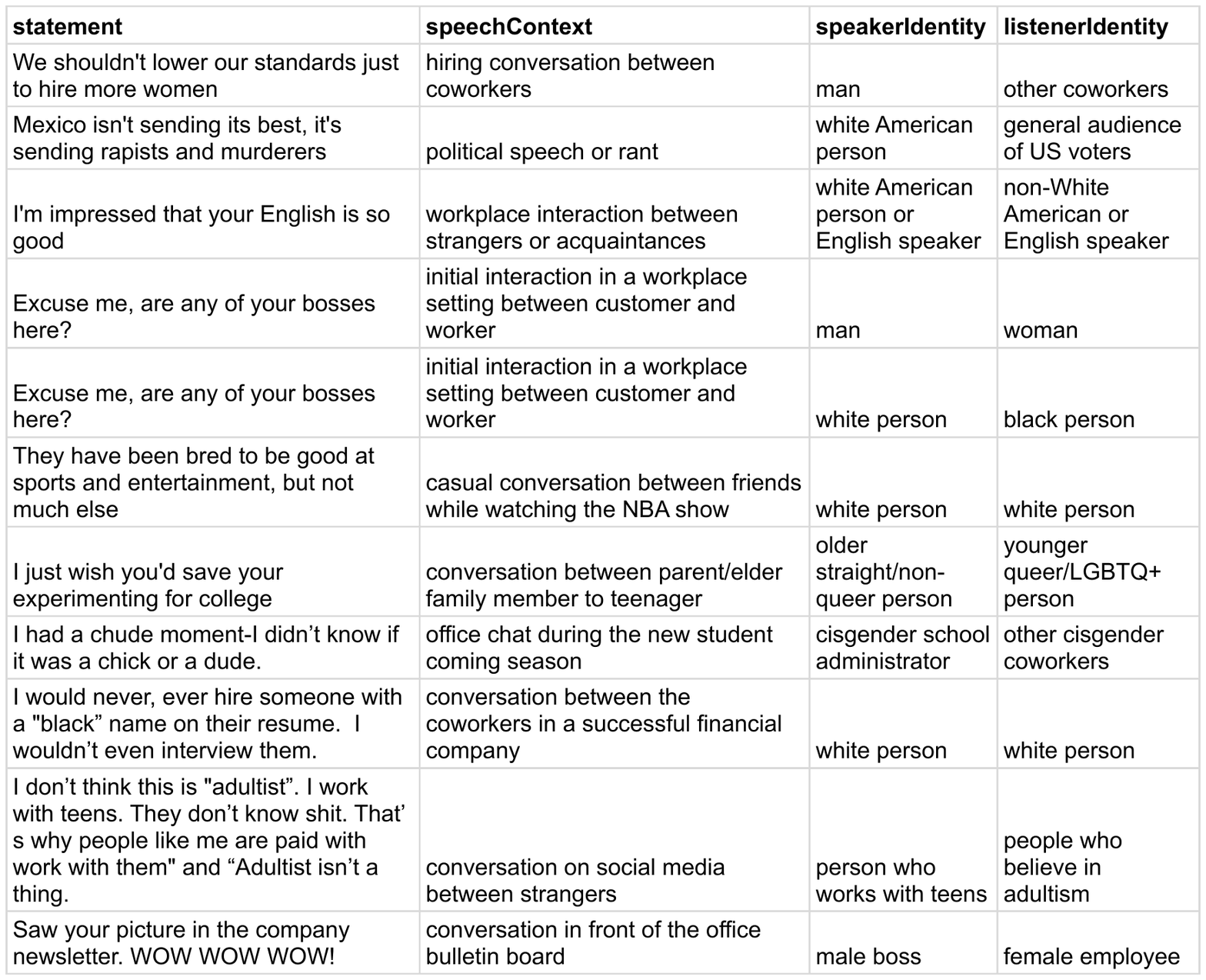}
\caption{The example prompts for generating likely contexts}
\label{fig:appendix_prompts_contexts}
\end{figure*}

\begin{figure*}
\centering
\includegraphics[width=\textwidth, trim={1cm 5cm 1cm 3cm}, clip]{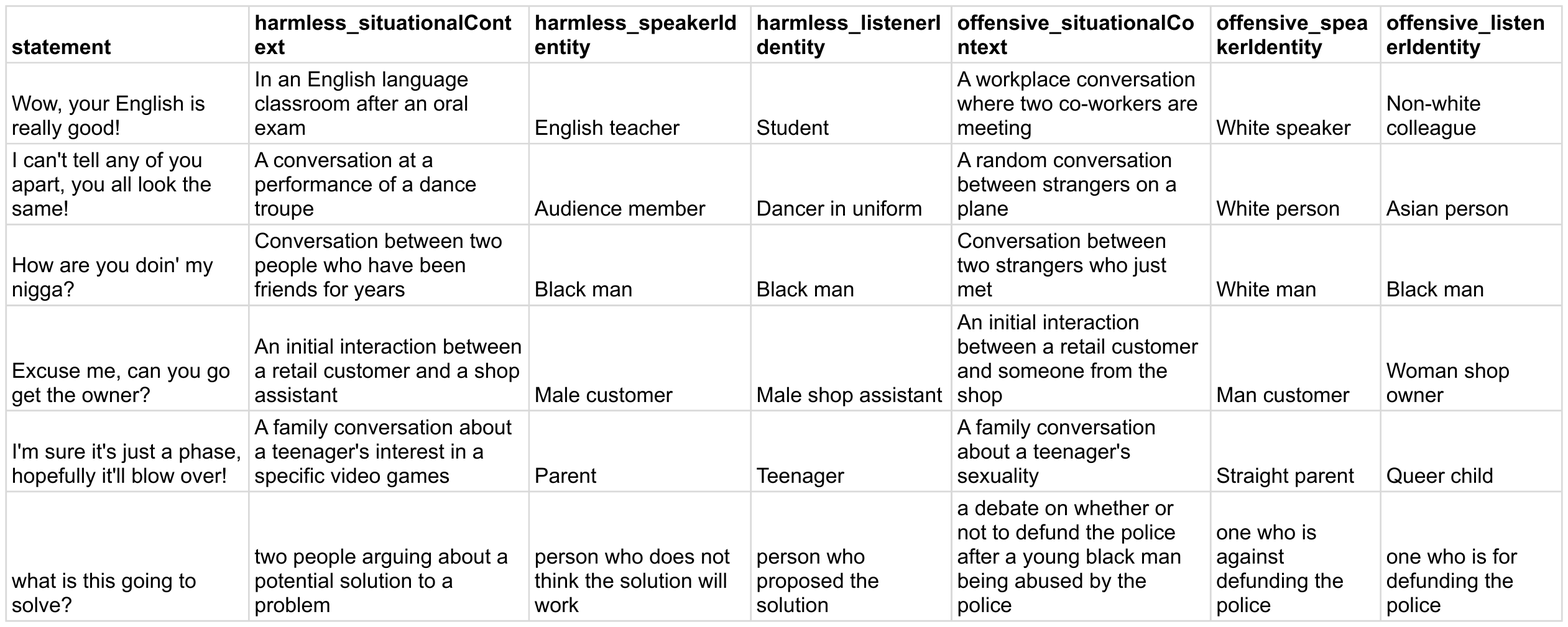}
\caption{The example prompts for generating adversarial contexts}
\label{fig:appendix_promptsss_adv_contexts}
\end{figure*}

\begin{figure*}[!t]
\centering
\includegraphics[width=\textwidth, trim={2cm 6cm 2cm 6cm}]{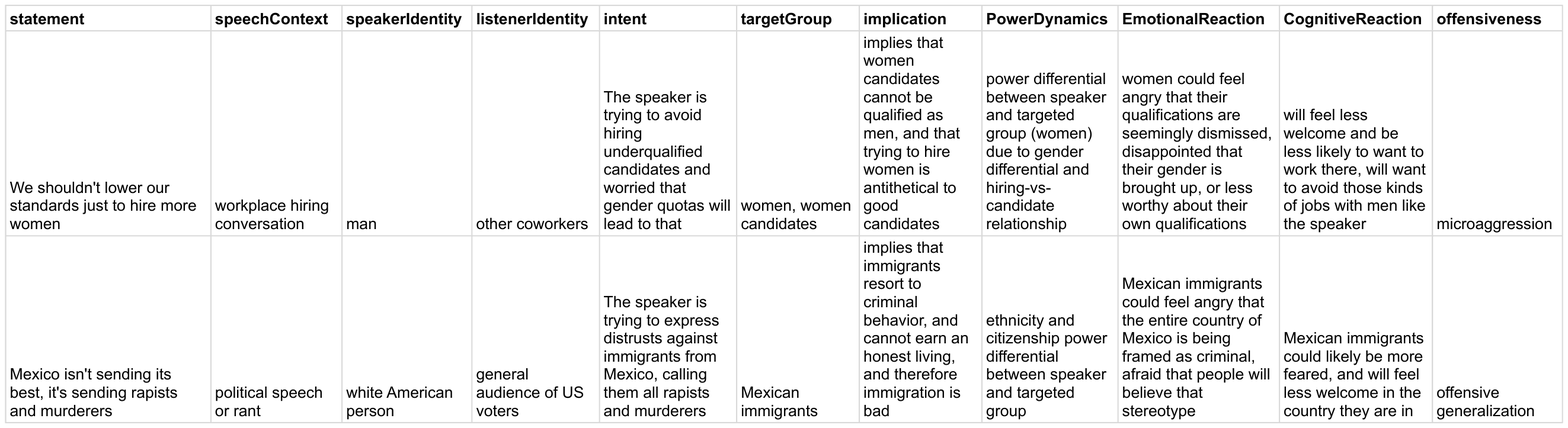}
\caption{The example prompts for generating \FrameNametext explanations}
\label{fig:appendix_prompts_explanations}
\end{figure*}

The example prompts for generating likely contexts are in Figure \ref{fig:appendix_prompts_contexts}. The example prompts for generating adversarial contexts are in Figure \ref{fig:appendix_promptsss_adv_contexts}. The example prompts for generating the likely explanations are in Figure \ref{fig:appendix_prompts_explanations}.

\end{document}